\documentclass{article}
\usepackage{iclr2023_conference,times}

\usepackage{hyperref}
\usepackage{url}

\usepackage{booktabs}
\usepackage{amsfonts}
\usepackage{amsmath}
\usepackage{mathtools}
\usepackage[capitalise,noabbrev]{cleveref}
\usepackage{tikz}
\usepackage{algorithm}
\usepackage{algorithmicx}
\usepackage[noend]{algpseudocode}
\usepackage{setspace}
\usepackage{subcaption}
\usepackage{wrapfig}
\usepackage{pifont}

\usepackage[T1]{fontenc}

\DeclarePairedDelimiterX{\DivX}[2]{(}{)}{#1\;\delimsize\|\;#2}
\newcommand{\kl}{D_\text{KL}\DivX}

\DeclarePairedDelimiter{\Paren}{\lparen}{\rparen}
\newcommand{\ent}[1][]{H_{#1}\Paren}
\newcommand{\sg}{\text{sg}\Paren}

\DeclarePairedDelimiter{\Brack}{\lbrack}{\rbrack}
\newcommand{\ev}[1][]{\mathbb{E}_{#1}\Brack}

\newcommand{\cond}{\:\lvert\:}

\DeclareMathOperator*{\sumt}{\textstyle\sum}

\newcommand{\obsparam}{\phi}
\newcommand{\dynparam}{\psi}
\newcommand{\actorparam}{\theta}
\newcommand{\criticparam}{\xi}
\newcommand{\obsprob}{p_\obsparam}
\newcommand{\dynprob}{p_\dynparam}
\newcommand{\dynfunc}{f_\dynparam}
\newcommand{\actor}{\pi_\actorparam}
\newcommand{\critic}{v_\criticparam}
\newcommand{\dataset}{\mathcal{D}}
\newcommand{\datasetsize}{T}
\newcommand{\datasettemp}{\tau}
\newcommand{\entropycoef}{\alpha_1}
\newcommand{\consistencycoef}{\alpha_2}

\newcommand{\rewardcoef}{\beta_1}
\newcommand{\discountcoef}{\beta_2}
\newcommand{\historylength}{\ell}
\newcommand{\wmbatchsize}{N}
\newcommand{\acbatchsize}{M}
\newcommand{\achorizon}{H}
\newcommand{\loss}{\mathcal{L}}
\newcommand{\actorentropycoef}{\eta}
\newcommand{\actorentropythreshold}{\Gamma}

\newcommand\keywordfont[1]{\textnormal{\textcolor{blue!90!black}{\ttfamily\bfseries #1}}\unskip}
\algrenewcommand\algorithmicend{{\keywordfont{end}}}
\algrenewcommand\algorithmicdo{{\keywordfont{do}}}
\algrenewcommand\algorithmicwhile{{\keywordfont{while}}}
\algrenewcommand\algorithmicfor{{\keywordfont{for}}}
\algrenewcommand\algorithmicforall{{\keywordfont{for all}}}
\algrenewcommand\algorithmicloop{{\keywordfont{loop}}}
\algrenewcommand\algorithmicrepeat{{\keywordfont{repeat}}}
\algrenewcommand\algorithmicuntil{{\keywordfont{until}}}
\algrenewcommand\algorithmicprocedure{{\keywordfont{procedure}}}
\algrenewcommand\algorithmicfunction{{\keywordfont{function}}}
\algrenewcommand\algorithmicif{{\keywordfont{if}}}
\algrenewcommand\algorithmicthen{{\keywordfont{then}}}
\algrenewcommand\algorithmicelse{{\keywordfont{else}}}
\algrenewcommand\algorithmicrequire{{\keywordfont{Require:}}}
\algrenewcommand\algorithmicensure{{\keywordfont{Ensure:}}}
\algrenewcommand\algorithmicreturn{{\keywordfont{return}}}
\algrenewcommand\textproc{}
\algnewcommand{\LineComment}[1]{\State \textcolor{green!50!black}{// #1}}

\title{Transformer-based World Models Are Happy With 100k Interactions}

\author{Jan Robine,\enspace Marc Höftmann,\enspace Tobias Uelwer,\enspace Stefan Harmeling \\
Department of Computer Science, Technical University of Dortmund, Germany\\
}

\iclrfinalcopy

\begin{document}

\maketitle

\begin{abstract}
Deep neural networks have been successful in many reinforcement learning
settings. However, compared to human learners they are overly data hungry. To
build a sample-efficient world model, we apply a transformer to real-world
episodes in an autoregressive manner: not only the compact latent states and the
taken actions but also the experienced or predicted rewards are fed into the
transformer, so that it can attend flexibly to all three modalities at different
time steps. The transformer allows our world model to access previous states
directly, instead of viewing them through a compressed recurrent state. By
utilizing the Transformer-XL architecture, it is able to learn long-term
dependencies while staying computationally efficient. Our transformer-based
world model (TWM) generates meaningful, new experience, which is used to train a
policy that outperforms previous model-free and model-based reinforcement
learning algorithms on the Atari 100k benchmark.
Our code is available at \url{https://github.com/jrobine/twm}. 
\end{abstract}

\section{Introduction}

\definecolor{ObservColor}{HTML}{9FA8DA}
\definecolor{RepresColor}{HTML}{64B5F6}
\definecolor{ActionColor}{HTML}{EF5350}
\definecolor{RewardColor}{HTML}{9CCC65}
\definecolor{DiscntColor}{HTML}{CE93D8}
\definecolor{TransformerColor}{HTML}{FFC107}
\definecolor{HiddenColor}{HTML}{FF9800}
\definecolor{VoidColor}{HTML}{BDBDBD}

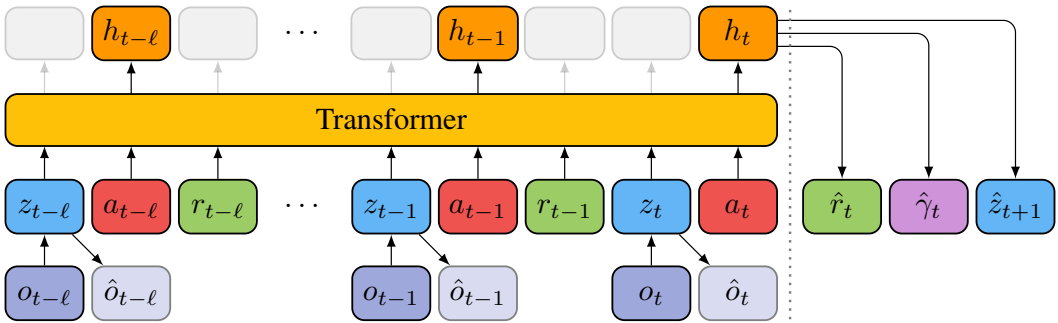
\begin{figure}[!b]
  \centering
  \resizebox{\textwidth}{!}{\begin{tikzpicture}
    \tikzset{Variable/.style={minimum width=0.9cm, minimum height=0.6cm, align=center, draw, semithick, rounded corners}}
    \tikzset{Observ/.style={Variable, black, fill=ObservColor}}
    \tikzset{Repres/.style={Variable, black, fill=RepresColor}}
    \tikzset{Action/.style={Variable, black, fill=ActionColor}}
    \tikzset{Reward/.style={Variable, black, fill=RewardColor}}
    \tikzset{Discnt/.style={Variable, black, fill=DiscntColor}}
    \tikzset{Transformer/.style={minimum height=0.6cm, align=center, text height=6pt, draw, semithick, rounded corners, black, fill=TransformerColor}}
    \tikzset{Hidden/.style={Variable, black, fill=HiddenColor}}
    \tikzset{Void/.style={Variable, black, fill=VoidColor, opacity=0.2}}
    \tikzset{Arrow/.style={->, >=latex, rounded corners}}
    \tikzset{VoidArrow/.style={->, >=latex, rounded corners, opacity=0.2}}

    \node[Observ] at (-7, -1) (otml) {$\strut{o}_{t-\historylength}$};
    \node[Repres] at (-7, 0) (ztml) {$\strut{z}_{t-\historylength}$};
    \node[Observ, draw=gray, fill=ObservColor!40!white] at (-6, -1) (recontml) {$\strut{\hat{o}}_{t-\historylength}$};
    \node[Void] at (-7, 2) (voidtml1) {\phantom{$h_{t-\historylength}$}};
    \node[Action] at (-6, 0) (atml) {$\strut{a}_{t-\historylength}$};
    \node[Hidden] at (-6, 2) (html) {$\strut{h}_{t-\historylength}$};
    \node[Reward] at (-5, 0) (rtml) {$\strut{r}_{t-\historylength}$};
    \node[Void] at (-5, 2) (voidtml2) {\phantom{$h_{t-\historylength}$}};
    \node at (-4, 0) (ellipsis1) {$\cdots$};
    \node at (-4, 2) (ellipsis2) {$\cdots$};
    \node[Observ] at (-3, -1) (otm1) {$\strut{o}_{t-1}$};
    \node[Repres] at (-3, 0) (ztm1) {$\strut{z}_{t-1}$};
    \node[Observ, draw=gray, fill=ObservColor!40!white] at (-2, -1) (recontm1) {$\strut{\hat{o}}_{t-1}$};
    \node[Void] at (-3, 2) (voidtm11) {\phantom{$h_{t-1}$}};
    \node[Action] at (-2, 0) (atm1) {$\strut{a}_{t-1}$};
    \node[Hidden] at (-2, 2) (htm1) {$\strut{h}_{t-1}$};
    \node[Reward] at (-1, 0) (rtm1) {$\strut{r}_{t-1}$};
    \node[Void] at (-1, 2) (voidtm12) {\phantom{$h_{t-1}$}};
    \node[Observ] at (0, -1) (ot) {$\strut{o}_t$};
    \node[Repres] at (0, 0) (zt) {$\strut{z}_t$};
    \node[Observ, draw=gray, fill=ObservColor!40!white] at (1, -1) (recont) {$\strut{\hat{o}}_t$};
    \node[Void] at (0, 2) (voidt) {\phantom{$h_t$}};
    \node[Action] at (1, 0) (at) {$\strut{a}_t$};
    \node[Hidden] at (1, 2) (ht) {$\strut{h}_t$};
    \node[Reward] at (2.2, 0) (rt) {$\strut{\hat{r}}_t$};
    \node[Discnt] at (3.2, 0) (dt) {$\strut\hat\gamma_t$};
    \node[Repres] at (4.2, 0) (ztp1) {$\strut{\hat{z}}_{t+1}$};

    \node[Transformer, minimum width=8.9cm] at (-3, 1) (transformer) {Transformer};

    \draw[Arrow] (otml.north) -- (ztml.south);
    \draw[Arrow] (ztml.north) -- ([yshift=0.38cm]ztml.north);
    \draw[Arrow] (atml.north) -- ([yshift=0.38cm]atml.north);
    \draw[Arrow] (rtml.north) -- ([yshift=0.38cm]rtml.north);
    \draw[Arrow] (otm1.north) -- (ztm1.south);
    \draw[Arrow] (ztm1.north) -- ([yshift=0.38cm]ztm1.north);
    \draw[Arrow] (atm1.north) -- ([yshift=0.38cm]atm1.north);
    \draw[Arrow] (rtm1.north) -- ([yshift=0.38cm]rtm1.north);
    \draw[Arrow] (ot.north) -- (zt.south);
    \draw[Arrow] (zt.north) -- ([yshift=0.38cm]zt.north);
    \draw[Arrow] (at.north) -- ([yshift=0.38cm]at.north);

    \draw[VoidArrow] ([yshift=-0.38cm]voidtml1.south) -- (voidtml1.south);
    \draw[Arrow] ([yshift=-0.38cm]html.south) -- (html.south);
    \draw[VoidArrow] ([yshift=-0.38cm]voidtml2.south) -- (voidtml2.south);
    \draw[VoidArrow] ([yshift=-0.38cm]voidtm11.south) -- (voidtm11.south);
    \draw[Arrow] ([yshift=-0.38cm]htm1.south) -- (htm1.south);
    \draw[VoidArrow] ([yshift=-0.38cm]voidtm12.south) -- (voidtm12.south);
    \draw[VoidArrow] ([yshift=-0.38cm]voidt.south) -- (voidt.south);
    \draw[Arrow] ([yshift=-0.38cm]ht.south) -- (ht.south);
    \draw[Arrow] ([yshift=-0.15cm]ht.east) -| (rt.north);
    \draw[Arrow] (ht.east) -| (dt.north);
    \draw[Arrow] ([yshift=0.15cm]ht.east) -| (ztp1.north);

    \draw[Arrow] (ztml) -- (recontml);
    \draw[Arrow] (ztm1) -- (recontm1);
    \draw[Arrow] (zt) -- (recont);

    \draw[gray, dotted, thick] (1.6, -1.3) -- (1.6, 2.3);
  \end{tikzpicture}}
  \vspace{-0.3cm}
  \caption{Our world model architecture. Observations $o_{t-\ell:t}$ are encoded
  using a CNN. Linear embeddings of stochastic, discrete latent states
  $z_{t-\ell:t}$, actions $a_{t-\ell:t}$, and rewards $r_{t-\ell:t}$ are fed
  into a transformer, which computes a deterministic hidden state $h_t$ at each
  time step. Predictions of the reward $r_t$, discount factor $\gamma_t$, and
  next latent state $z_{t+1}$ are computed based on $h_t$ using MLPs.}
  \label{fig:transformer}
\end{figure}

Deep reinforcement learning methods have shown great success on many challenging
decision making problems. Notable methods include DQN \citep{dqn}, PPO
\citep{ppo}, and MuZero \citep{muzero}. However, most algorithms require
hundreds of millions of interactions with the environment, whereas humans often
can achieve similar results with less than 1\% of these interactions, i.e., they
are more sample-efficient. The large amount of data that is necessary renders a
lot of potential real world applications of reinforcement learning impossible.

Recent works have made a lot of progress in advancing the sample efficiency of
RL algorithms: model-free methods have been improved with auxiliary objectives
\citep{curl}, data augmentation (\citealp{drq}, \citealp{rad}), or both
\citep{spr}. Model-based methods have been successfully applied to complex
image-based environments and have either been used for planning, such as
EfficientZero \citep{efficient-zero}, or for learning behaviors in imagination,
such as SimPLe \citep{simple}.

A promising model-based concept is learning in imagination
(\citealp{world-models}; \citealp{simple}; \citealp{dreamerv1};
\citealp{dreamerv2}): instead of learning behaviors from the collected
experience directly, a generative model of the environment dynamics is learned
in a \mbox{(self-)supervised} manner. Such a so-called \textit{world model} can
create new trajectories by iteratively predicting the next state and reward.
This allows for potentially indefinite training data for the reinforcement
learning algorithm without further interaction with the real environment. A
world model might be able to generalize to new, unseen situations, because of
the nature of deep neural networks, which has the potential to drastically
increase the sample efficiency. This can be illustrated by a simple example: in
the game of Pong, the paddles and the ball move independently. In the best case,
a successfully trained world model would imagine trajectories with paddle and
ball configurations that have never been observed before, which enables learning
of improved behaviors.

In this paper, we propose to model the world with transformers
\citep{transformer}, which have significantly advanced the field of natural
language processing and have been successfully applied to computer vision tasks
\citep{vit}. A transformer is a sequence model consisting of multiple
self-attention layers with residual connections. In each self-attention layer
the inputs are mapped to keys, queries, and values. The outputs are computed by
weighting the values by the similarity of keys and queries. Combined with causal
masking, which prevents the self-attention layers from accessing future time
steps in the training sequence, transformers can be used as autoregressive
generative models. The Transformer-XL architecture \citep{transformer-xl} is
much more computationally efficient than vanilla transformers at inference time
and introduces relative positional encodings, which remove the dependence on
absolute time steps.

\vspace{-0.3\baselineskip}
\paragraph{Our contributions:}
The contributions of this work can be summarized as follows:
\begin{enumerate}
\item We present a new autoregressive world model based on the Transformer-XL
  \citep{transformer-xl} architecture and a model-free agent trained in latent
  imagination. Running our policy is computationally efficient, as the
  transformer is not needed at inference time. This is in contrast to related
  works \citep{dreamerv1,dreamerv2,transdreamer} that require the full world
  model during inference.
\item Our world model is provided with information on how much reward has
  already been emitted by feeding back predicted rewards into the world model.
  As shown in our ablation study, this improves performance.
\item We rewrite the balanced KL divergence loss of \citet{dreamerv2} to allow
  us to fine-tune the relative weight of the involved entropy and cross-entropy
  terms.
\item We introduce a new thresholded entropy loss that stabilizes the policy's
  entropy during training and hereby simplifies the selection of hyperparameters
  that behave well across different games.
\item We propose a new effective sampling procedure for the growing dataset of
  experience, which balances the training distribution to shift the focus
  towards the latest experience. We demonstrate the efficacy of this procedure
  with an ablation study.
\item We compare our transformer-based world model (TWM) on the Atari 100k
  benchmark with recent sample-efficient methods and obtain excellent results.
  Moreover, we report empirical confidence intervals of the aggregate metrics as
  suggested by \citet{aggregate}.
\end{enumerate}

\section{Method}
We consider a partially observable Markov decision process (POMDP) with discrete
time steps \mbox{$t \in \mathbb{N}$}, scalar rewards \mbox{$r_t \in
\mathbb{R}$}, high-dimensional image observations  \mbox{$o_t \in \mathbb{R}^{h
\times w \times c}$}, and discrete actions \mbox{$a_t \in \{1, \ldots, m\}$},
which are generated by some policy \mbox{$a_t \sim \pi(a_t \cond o_{1:t},
a_{1:t-1})$}, where  $o_{1:t}$ and $a_{1:t-1}$ denote the sequences of
observations and actions up to time steps $t$ and \mbox{$t-1$}, respectively.
Episode ends are indicated by a boolean variable \mbox{$d_t \in \{0, 1\}$}.
Observations, rewards, and episode ends are jointly generated by the unknown
environment dynamics \mbox{$o_t, r_t,d_t \sim p(o_t,r_t,d_t \cond o_{1:t-1},
a_{1:t-1})$}. The goal is to find a policy $\pi$ that maximizes the expected sum
of discounted rewards \mbox{$\ev[\pi\!][\big]{\sum_{t=1}^\infty \gamma^{t-1}
r_t}\!$}, where \mbox{$\gamma \in [0,1)$} is the discount factor. Learning in
imagination consists of three steps that are repeated iteratively: learning the
dynamics, learning a policy, and interacting in the real environment. In this
section, we describe our world model and policy, concluding with the training
procedure.

\subsection{World Model}
Our world model consists of an observation model and a dynamics model, which do
not share parameters. \cref{fig:transformer} illustrates our combined world
model architecture.

\paragraph{Observation Model:}
The observation model is a variational autoencoder \citep{vae}, which encodes
observations $o_t$ into compact, stochastic latent states $z_t$ and reconstructs
the observations with a decoder, which in our case is only required to obtain a
learning signal for $z_t$:
\begingroup
\addtolength{\jot}{0.1em}
\begin{equation}
\begin{alignedat}{3}
  & \text{Observation encoder:} \quad & z_t & \sim \obsprob(z_t \cond o_t) \\
  & \text{Observation decoder:} \quad & \hat{o}_t & \sim \obsprob(\hat{o}_t \cond z_t).
\end{alignedat}
\end{equation}
\endgroup We adopt the neural network architecture of DreamerV2
\citep{dreamerv2} with slight modifications for our observation model. Thus, a
latent state $z_t$ is discrete and consists of a vector of $32$ categorical
variables with $32$ categories. The observation decoder reconstructs the
observation and predicts the means of independent standard normal distributions
for all pixels. The role of the observation model is to capture only
non-temporal information about the current time step, which is different from
\citet{dreamerv2}. However, we include short-time temporal information, since a
single observation $o_t$ consists of four frames (aka frame stacking, see also
\cref{sec:policy-input}).

\paragraph{Autoregressive Dynamics Model:} \label{sec:dynamics-model} The
dynamics model predicts the next time step conditioned on the history of its
past predictions. The backbone is a deterministic aggregation model $\dynfunc$
which computes a deterministic hidden state $h_t$ based on the history of the
$\historylength$ previously generated latent states, actions, and rewards.
Predictors for the reward, discount, and next latent state are conditioned on
the hidden state. The dynamics model consists of these components:
\begingroup
\addtolength{\jot}{0.1em}
\begin{equation}
\begin{alignedat}{3}
  & \text{Aggregation model:}         &  \quad h_t & = \dynfunc(z_{t-\historylength:t},a_{t-\historylength:t},r_{t-\historylength:t-1}) \\
  & \text{Reward predictor:}          &  \quad \hat{r}_t & \sim \dynprob(\hat{r}_t \cond h_t) \\
  & \text{Discount predictor:}        &  \quad \hat{\gamma}_t & \sim \dynprob(\hat{\gamma}_t \cond h_t) \\
  & \text{Latent state predictor:}\ \ &  \quad \hat{z}_{t+1} & \sim \dynprob(\hat{z}_{t+1} \cond h_t).
\end{alignedat}
\end{equation}
\endgroup
The aggregation model is implemented as a causally masked Transformer-XL
\citep{transformer-xl}, which enhances vanilla transformers \citep{transformer}
with a recurrence mechanism and relative positional encodings. With these
encodings, our world model learns the dynamics independent of absolute time
steps. Following \citet{decision-transformer}, the latent states, actions, and
rewards are sent into modality-specific linear embeddings before being passed to
the transformer. The number of input tokens is $3\historylength - 1$, because of
the three modalities (latent states, actions, rewards) and the last reward not
being part of the input. We consider the outputs of the action modality as the
hidden states and disregard the outputs of the other two modalities (see
\cref{fig:transformer}; orange boxes vs. gray boxes).

The latent state, reward, and discount predictors are implemented as multilayer
perceptrons (MLPs) and compute the parameters of a vector of independent
categorical distributions, a normal distribution, and a Bernoulli distribution,
respectively, conditioned on the deterministic hidden state. The next state is
determined by sampling from $\dynprob(\hat{z}_{t+1} \cond h_t)$. The reward and
discount are determined by the mean of $\dynprob(\hat{r}_t \cond h_t)$ and
$\dynprob(\hat{\gamma}_t \cond h_t)$, respectively.

As a consequence of these design choices, our world model has the following
beneficial properties:
\begin{enumerate}
  \item The dynamics model is autoregressive and has direct access to its
    previous outputs.
  \item Training is efficient since sequences are processed in parallel
    (compared with RNNs).
  \item Inference is efficient because outputs are cached (compared with
    vanilla Transformers).
  \item Long-term dependencies can be captured by the recurrence mechanism.
\end{enumerate}

We want to provide an intuition on why a fully autoregressive dynamics model is
favorable: First, the direct access to previous latent states enables to model
more complex dependencies between them, compared with RNNs, which only see them
indirectly through a compressed recurrent state. This also has the potential to
make inference more robust, since degenerate predictions can be ignored more
easily. Second, because the model sees which rewards it has produced previously,
it can react to its own predictions. This is even more significant when the
rewards are sampled from a probability distribution, since the introduced noise
cannot be observed without autoregression.

\paragraph{Loss Functions:}
The observation model can be interpreted as a variational autoencoder with a
temporal prior, which is provided by the latent state predictor. The goal is to
keep the distributions of the encoder and the latent state predictor close to
each other, while slowly adapting to new observations and dynamics.
\citet{dreamerv2} apply a balanced KL divergence loss, which lets them control
which of the two distributions should be penalized more.  To control the
influences of its subterms more precisely, we disentangle this loss and obtain a
\textit{balanced cross-entropy loss} that computes the cross-entropy
$\ent{\obsprob(z_{t+1} \cond o_{t+1}),\dynprob(\hat{z}_{t+1} \cond h_t)}$ and
the entropy $\ent{\obsprob(z_t \cond o_t)}$ explicitly. Our derivation can be
found in \cref{sec:loss-derivation}. We call the cross-entropy term for the
observation model the \textit{consistency loss}, as its purpose is to prevent
the encoder from diverging from the dynamics model. The entropy regularizes the
latent states and prevents them from collapsing to one-hot distributions. The
observation decoder is optimized via negative log-likelihood, which provides a
rich learning signal for the latent states. In summary, we optimize a
self-supervised loss function for the observation model that is the expected sum
over the decoder loss, the entropy regularizer and the consistency loss
\begin{equation} \label{eq:observation-loss}
  \loss_\obsparam^\text{Obs.} = \ev[\!][\bigg]{\,\sumt_{t=1}^T -\underbrace{\ln \obsprob(o_t \cond z_t)}_\text{decoder} - \underbrace{\entropycoef \ent{\obsprob(z_t \cond o_t)}}_\text{entropy regularizer} + \underbrace{\consistencycoef \ent{\obsprob(z_t \cond o_t),\dynprob(\hat{z}_t \cond h_{t-1})}}_\text{consistency}}\!,
\end{equation}
where the hyperparameters $\entropycoef, \consistencycoef \ge 0$ control the
relative weights of the terms.

For the balanced cross-entropy loss, we also minimize the cross-entropy in the
loss of the dynamics model, which is how we train the latent state predictor.
The reward and discount predictors are optimized via negative log-likelihood.
This leads to a self-supervised loss for the dynamics model
\begin{equation} \label{eq:dynamics-loss}
  \loss_\dynparam^\text{Dyn.} = \ev[\!][\bigg]{\,\sumt_{t=1}^T \underbrace{\ent{\obsprob(z_{t+1} \cond o_{t+1}),\dynprob(\hat{z}_{t+1} \cond h_t)}}_\text{latent state predictor} - \underbrace{\rewardcoef \ln \dynprob(r_t \cond h_t)}_\text{reward predictor} - \underbrace{\discountcoef \ln \dynprob(\gamma_t \cond h_t)}_\text{discount predictor}\,}\!,
\end{equation}
with coefficients $\rewardcoef, \discountcoef \ge 0$ and where $\gamma_t = 0$
for episode ends ($d_t = 1$) and $\gamma_t = \gamma$ otherwise.

\subsection{Policy} \label{sec:policy}
Our policy $\pi_\theta(a_t \cond \hat{z}_t)$ is trained on imagined trajectories
using a mainly standard advantage actor-critic \citep{a3c} approach. We train
two separate networks: an actor $a_t \sim \actor(a_t \cond \hat{z}_t)$ with
parameters $\actorparam$ and a critic $\critic(\hat{z}_t)$ with parameters
$\criticparam$. We compute the advantages via Generalized Advantage Estimation
\citep{gae} while using the discount factors predicted by the world model
$\hat{\gamma}_t$ instead of a fixed discount factor for all time steps. As in
DreamerV2 \citep{dreamerv2}, we weight the losses of the actor and the critic by
the cumulative product of the discount factors, in order to softly account for
episode ends.

\paragraph{Thresholded Entropy Loss:} \label{sec:entropy-threshold} We penalize
the objective of the actor with a slightly modified version of the usual entropy
regularization term \citep{a3c}. Our penalty normalizes the entropy and only
takes effect when the entropy falls below a certain threshold
\begin{equation}
  \mathcal{L}^\text{Ent.}_\theta =\max\!\left(0, \actorentropythreshold - \frac{H(\pi_\theta)}{\ln(m)}\right)\hspace{-3pt},
\end{equation}
where $0\le \Gamma\le 1$ is the threshold hyperparameter, $H(\pi_\theta)$ is the
entropy of the policy, $m$ is the number of discrete actions, and $\ln(m)$ is
the maximum possible entropy of the categorical action distribution. By doing
this, we explicitly control the percentage of entropy that should be preserved
across all games independent of the number of actions. This ensures exploration
in the real environment and in imagination without the need for
$\epsilon$-greedy action selection or changing the temperature of the action
distribution. We also use the same stochastic policy when evaluating our agent
in the experiments. The idea of applying a hinge loss to the entropy was first
introduced by \citet{hinge-entropy} in the context of supervised learning. In
\cref{sec:additional-ablation-studies} we show the effect of this loss.

\paragraph{Choice of Policy Input:}
\label{sec:policy-input}
The policy computes an action distribution $\pi_\theta(a_t \cond x_t)$ given
some view $x_t$ of the state. For instance, $x_t$ could be $o_t$, $z_t$, or
$[z_t,h_t]$ at inference time, i.e., when applied to the real environment, or
the respective predictions of the world model $\hat{o}_t$, $\hat{z}_t$, or
$[\hat{z}_t,h_t]$ at training time. This view has to be chosen carefully, since
it can have a significant impact on the performance of the policy and it affects
the design choices for the world model. Using $x_t = o_t$ (or $\hat{o}_t$) is
relatively stable even with imperfect reconstructions $\hat{o}_t$, as the
underlying distribution of observations $p(o)$ does not change during training.
However, it is also less computationally efficient, since it requires
reconstructing the observations during imagination and additional convolutional
layers for the policy. Using $x_t = z_t$ (or $\hat{z}_t$) is slightly less
stable, as the policy has to adopt to the changes of the distributions
$\obsprob(z_t \cond o_t)$ and $\dynprob(\hat{z}_{t+1} \cond h_t)$ during
training. Nevertheless, the entropy regularizer and consistency loss in
\cref{eq:observation-loss} stabilize these distributions. Using \mbox{$x_t =
[z_t,h_t]$} (or $[\hat{z}_t,h_t]$) provides the agent with a summary of the
history of experience, but it also adds the burden of running the transformer at
inference time. Model-free agents already perform well on most Atari games when
using a stack of the most recent frames (e.g., \citealt{dqn,ppo}). Therefore, we
choose \mbox{$x_t = z_t$} and apply frame stacking at inference time in order to
incorporate short-time information directly into the latent states. At training
time we use $x_t = \hat{z}_t$, i.e., the predicted latent states, meaning no
frame stacking is applied. As a consequence, our policy is computationally
efficient at training time (no reconstructions during imagination) and at
inference time (no transformer when running in the real environment).

\subsection{Training}
As is usual for learning with world models, we repeatedly (i) collect experience
in the real environment with the current policy, (ii) improve the world model
using the past experience, (iii) improve the policy using new experience generated
by the world model.

During training we build a dataset $\dataset = [(o_1, a_1, r_1, d_1), \ldots,
(o_\datasetsize, a_\datasetsize, r_\datasetsize, d_\datasetsize)]$ of the
collected experience. After collecting new experience with the current policy,
we improve the world model by sampling $\wmbatchsize$ sequences of length
$\historylength$ from $\dataset$ and optimizing the loss functions in
\cref{eq:observation-loss,eq:dynamics-loss} using stochastic gradient descent.
After performing a world model update, we select $\acbatchsize$ of the
$\wmbatchsize \times \historylength$ observations and encode them into latent
states to serve as initial states for new trajectories. The dynamics model
iteratively generates these $\acbatchsize$ trajectories of length $\achorizon$
based on actions provided by the policy. Subsequently, the policy is improved
with standard model-free objectives, as described in \cref{sec:policy}. In
\cref{algo:main} we present pseudocode for training the world model and the
policy.

\begin{figure}
  \centering
  \includegraphics[height=3.5cm]{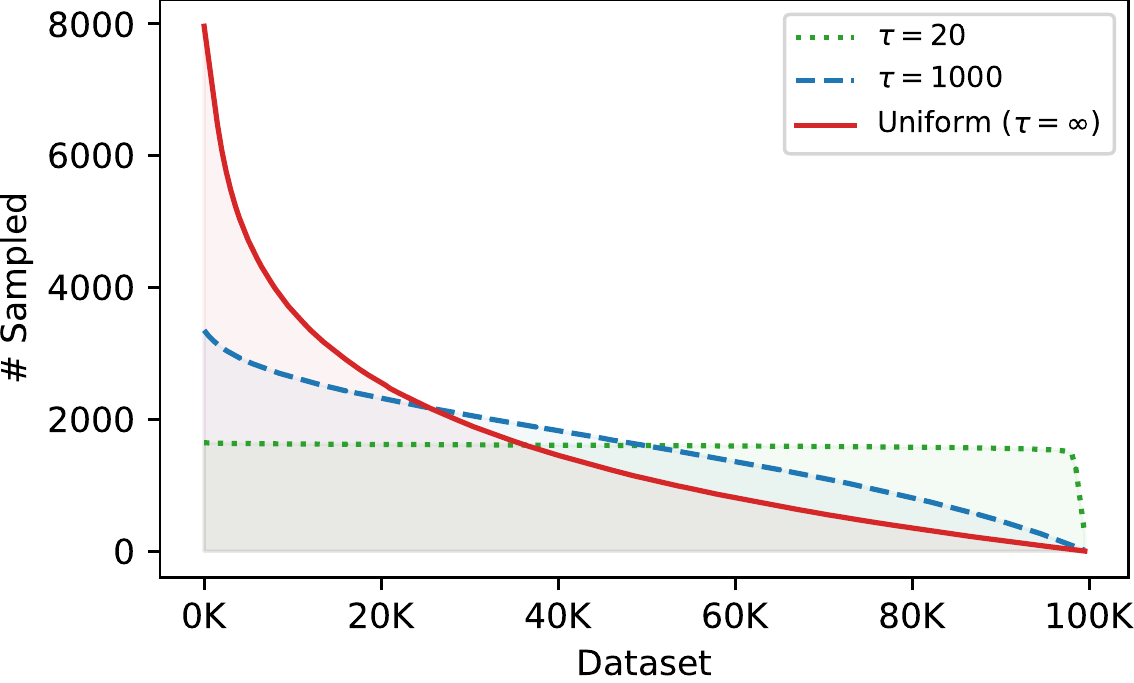}
  \hspace{0.2cm}
  \includegraphics[height=3.5cm]{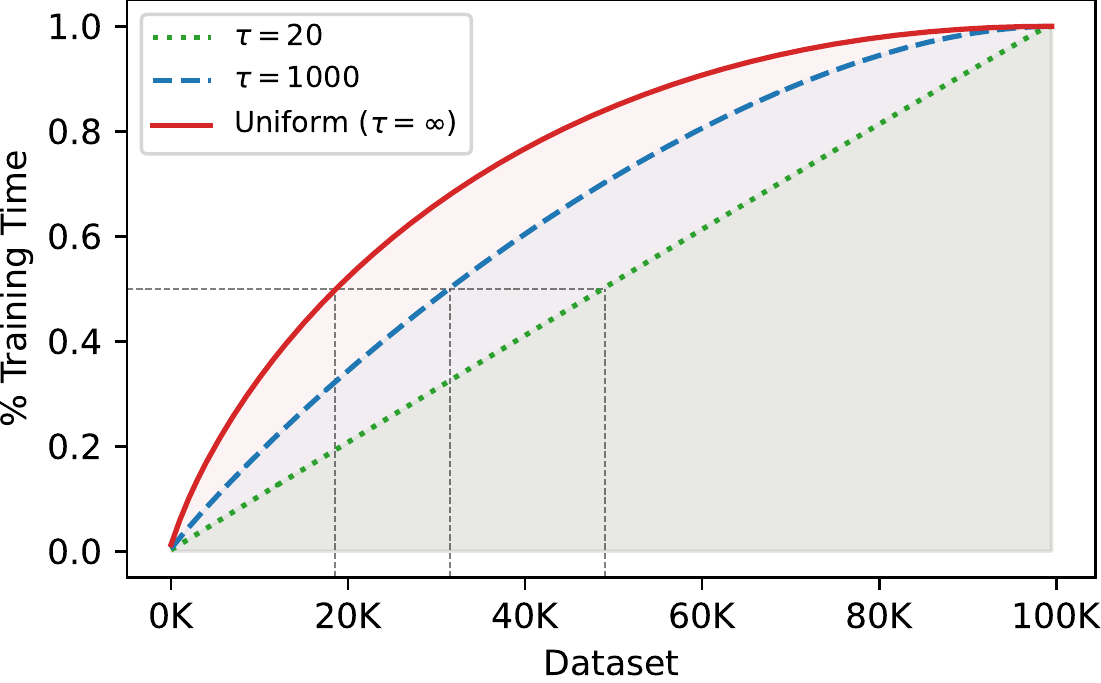}
  \caption{Comparing our balanced dataset sampling procedure (see
    \cref{eq:sample-prob}) for different values of $\tau$ with uniform sampling ($\tau=\infty$). The
    x-axes correspond to the entries in dataset $\dataset$ in the
    order they are experienced. The left plot shows the number of
    times an entry has been selected for training the world model. The
    right plot shows the relative amount of training time that has
    been spent on the data up to that entry. E.g., with uniform
    sampling, $50\%$ of the training time is used for the first $19$K
    entries, whereas for $\tau = 20$ approximately the same
    time is spend on both halves of the dataset.}
  \label{fig:sampling}
\end{figure}

\paragraph{Balanced Dataset Sampling:} \label{sec:sampling} Since the dataset
grows slowly during training, uniform sampling of trajectories focuses too
heavily on early experience, which can lead to overfitting especially in the low
data regime. Therefore, we keep visitation counts $v_1, \ldots, v_\datasetsize$,
which are incremented every time an entry is sampled as start of a sequence.
These counts are converted to probabilities using the softmax function
\begin{equation} \label{eq:sample-prob}
  (p_1, \ldots, p_T) = \mathrm{softmax}\left(-\tfrac{v_1}{\datasettemp}, \dots, {-\tfrac{v_\datasetsize}{\datasettemp}}\right)\hspace{-2.5pt},
\end{equation}
where $\datasettemp>0$ is a temperature hyperparameter. With our sampling
procedure, new entries in the dataset are oversampled and are selected more
often than old ones. Setting $\datasettemp = \infty$ restores uniform sampling
as a special case, whereas reducing $\datasettemp$ increases the amount of
oversampling. See \cref{fig:sampling} for a comparison. We empirically show the
effectiveness in \cref{sec:ablation-uniform-sampling}.

\section{Experiments}

To compare data-efficient reinforcement learning algorithms, \citet{simple}
proposed the Atari 100k benchmark, which uses a subset of $26$ Atari games from
the Arcade Learning Environment \citep{ale} and limits the number of
interactions per game to $100$K.  This corresponds to $400$K frames (because of
frame skipping) or roughly $2$ hours of gameplay, which is $500$ times less than
the usual $200$ million frames (e.g., \citealt{dqn,ppo,dreamerv2}).

We compare our method with five strong competitors on the Atari 100k benchmark:
(i) SimPLe \citep{simple} implements a world model as an action-conditional
video prediction model and trains a policy with PPO \citep{ppo}, (ii) DER
\citep{data-efficient-rainbow} is a variant of Rainbow \citep{rainbow}
fine-tuned for sample efficiency, (iii) CURL \citep{curl} improves representations
using contrastive learning as an auxiliary task and is combined with DER, (iv)
DrQ \citep{drq} improves DQN by averaging Q-value estimates over multiple data
augmentations of observations, and (v) SPR \citep{spr} forces representations to
be consistent across multiple time steps and data augmentations by extending
Rainbow with a self-supervised consistency loss.

\subsection{Results}

\begin{figure}
  \centering
  \includegraphics[width=\textwidth]{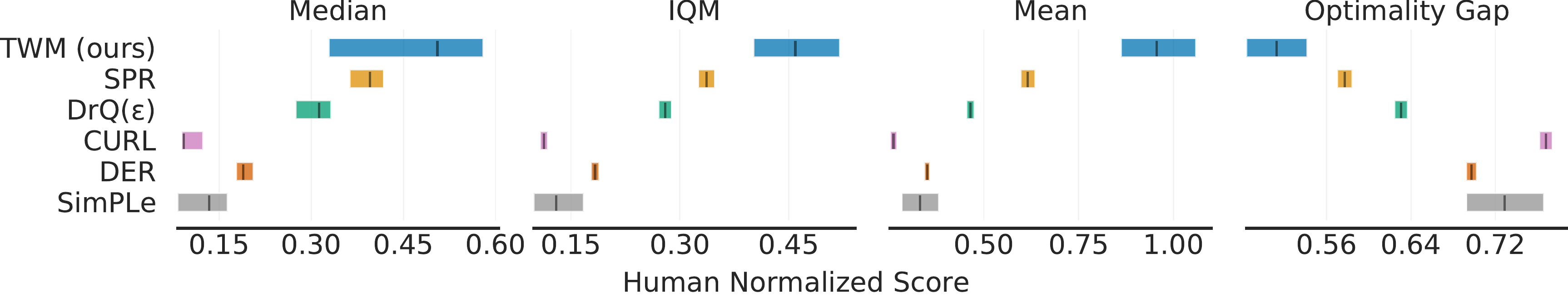}
  \vspace{-0.5cm}
  \caption{Aggregate metrics on the Atari 100k benchmark with $95\%$ stratified
  bootstrap confidence intervals \citep{aggregate}. Higher median, interquartile
  mean (IQM), and mean, but lower optimality gap indicate better performance.
  Scores for previous methods are from \citet{aggregate} with $100$ runs per
  game (except SimPLe with $5$ runs). We evaluate $5$ runs per game, leading to
  wider confidence intervals.}
  \label{fig:interval-estimates}
\end{figure}

\begin{wrapfigure}{r}{0.45\textwidth}
  \centering
  \vspace{-1.1\baselineskip}
  \includegraphics[width=0.4\textwidth]{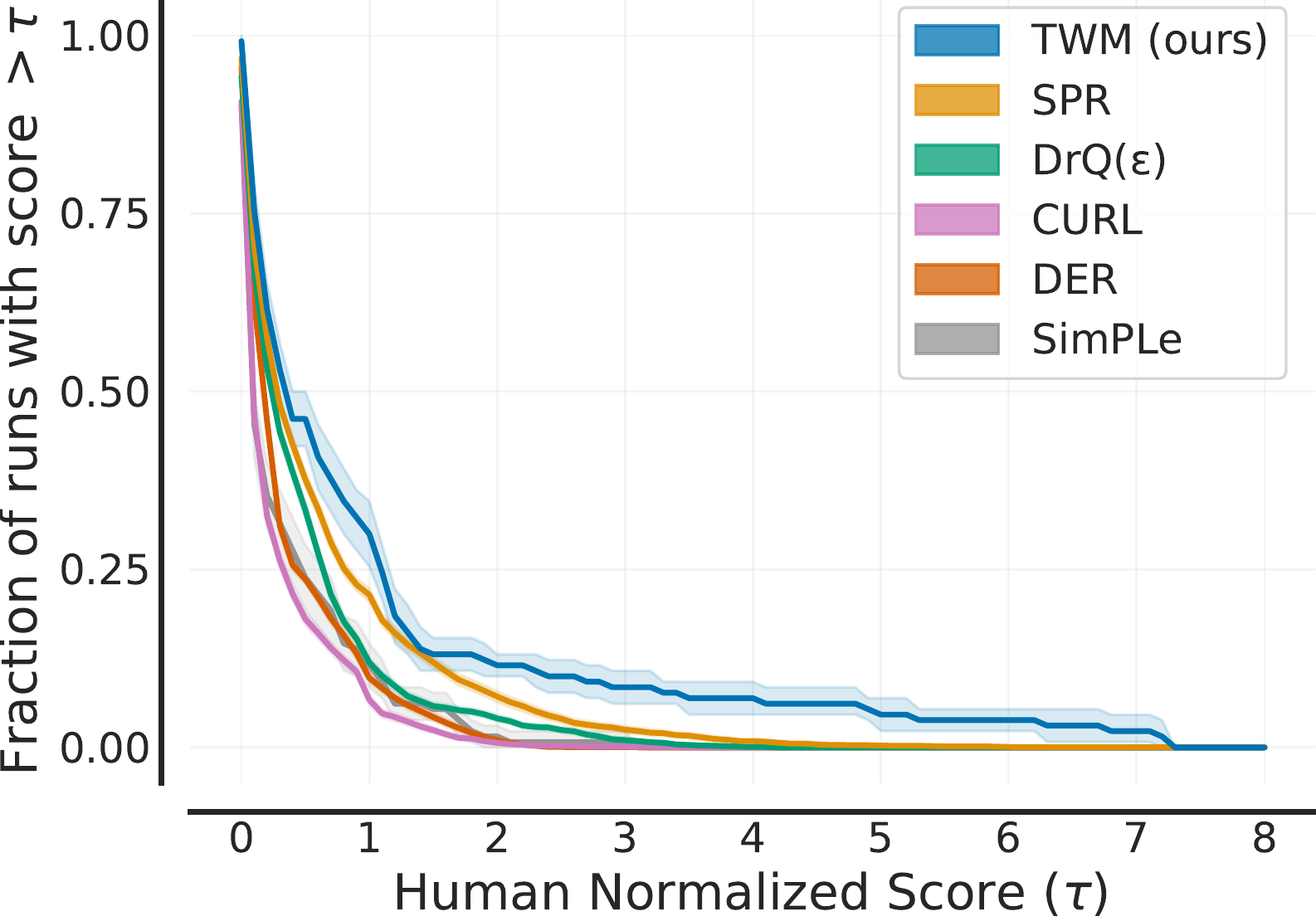}
  \caption{Performance profiles on the Atari 100k benchmark based on score
  distributions \citep{aggregate}. It shows the fraction of runs across all
  games (y-axis) above a human normalized score (x-axis). Shaded regions show
  pointwise $95\%$ confidence bands.}
  \label{fig:performance-profiles}
  \vspace{-\baselineskip}
\end{wrapfigure}

We follow the advice of \citet{aggregate} who found significant discrepancies
between reported point estimates of mean (and median) scores and a thorough
statistical analysis that includes statistical uncertainty. Thus, we report
confidence interval estimates of the aggregate metrics median, interquartile
mean (IQM), mean, and optimality gap in \cref{fig:interval-estimates} and
performance profiles in \cref{fig:performance-profiles}, which we created using
the toolbox provided by \citet{aggregate}. The metrics are computed on human
normalized scores, which are calculated as
\texttt{(score\_agent\,-\,score\_random)/}
\texttt{(score\_human\,-\,score\_random)}. We report the unnormalized scores per
game in \cref{tab:results}. We compare with new scores for DER, CURL, DrQ, and
SPR that were evaluated on $100$ runs and provided by \citet{aggregate}. They
report scores for the improved DrQ($\varepsilon$), which is DrQ evaluated with
standard $\varepsilon$-greedy parameters.
We perform $5$ runs per game and compute the average score over $100$ episodes
at the end of training for each run. TWM shows a significant improvement over
previous approaches in all four aggregate metrics and brings the optimality gap
closer to zero.

\subsection{Analysis}

\begin{figure}
  \captionsetup[subfigure]{aboveskip=3pt,belowskip=0pt}

  \begin{subfigure}{\textwidth}
    \includegraphics[width=\textwidth,trim=3247 60 5405 18,clip]{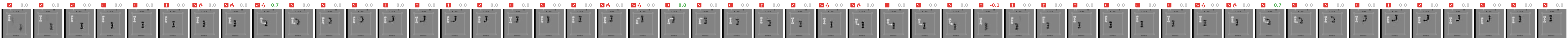}
    \caption{Boxing. The player (white) presses \textit{fire}, hits the
      opponent, and gets a reward.}
    \vspace{0.1cm}
  \end{subfigure}
  \begin{subfigure}{\textwidth}
    \includegraphics[width=\textwidth,trim=4110 60 4540 18,clip]{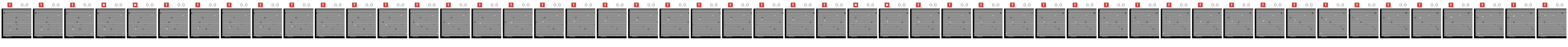}
    \caption{Freeway. The player moves up and bumps into a car. The world model
      correctly pushes the player down, although \textit{up} is still pressed.
      The movement of the cars is modeled correctly.}
    \label{fig:trajectory-freeway}
    \vspace{-0.2\baselineskip}
  \end{subfigure}
  \caption{Trajectories imagined by our world model. Above each frame we show
    the performed action and the produced reward.}
  \label{fig:trajectories}
\end{figure}

\begin{figure}
  \centering
  \includegraphics[width=0.75\textwidth,trim=8 8 6 8]{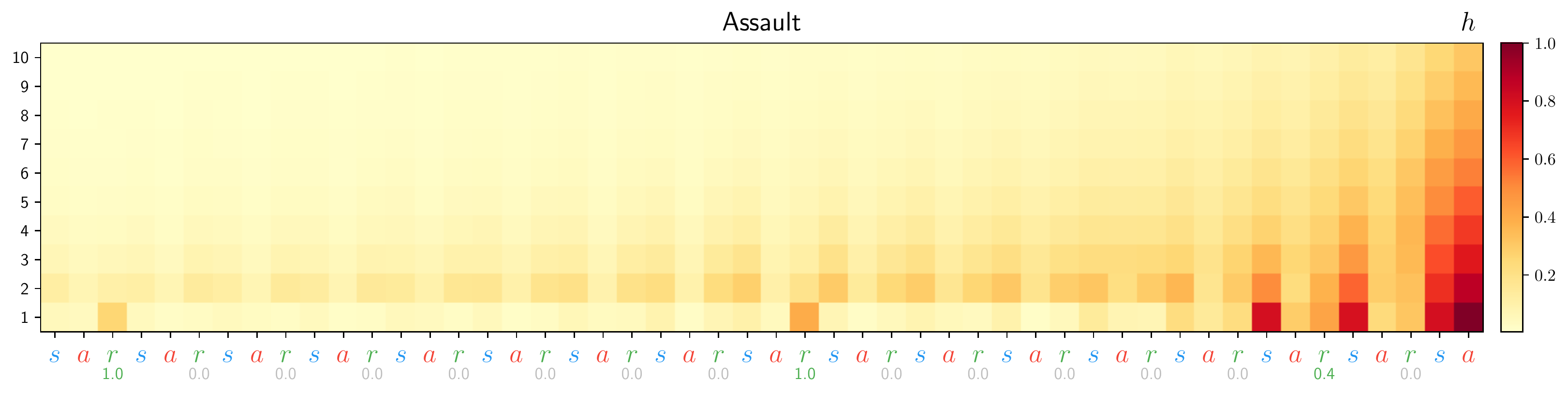}
  \caption{Attention map of the learned transformer for the current hidden state
    $h$, computed on an imagined trajectory for the game Assault. The x-axis
    corresponds to the input sequence with the three modalities (states,
    actions, rewards), where the two rightmost columns are the current state and
    action. The y-axis corresponds to the layer of the transformer.}
  \label{fig:attention-map}
\end{figure}

In \cref{fig:trajectories} we show imagined trajectories of our world model. In
\cref{fig:attention-map} we visualize an attention map of the transformer for an
imagined sequence. In this example a lot of weight is put on the current action
and the last three states. However, the transformer also attends to states and
rewards in the past, with only past actions being mostly ignored. The two high
positive rewards also get high attention, which confirms that the rewards in the
input sequence are used by the world model. We hypothesize that these rewards
correspond to some events that happened in the environment and this information
can be useful for prediction.

An extended analysis can be found in \cref{sec:additional-analysis}, including
more imagined trajectories and attention maps (and a description of the
generation of the plots), sample efficiency, stochasticity of the world model,
long sequence imagination, and frame stacking.

\begin{table}
  \centering
  \caption{Mean scores on the Atari 100k benchmark per game as well as the
  aggregated human normalized mean and median. We perform $5$ runs per game and
  compute the average over $100$ episodes at the end of training for each run.
  Bold numbers indicate the best scores.}
  \label{tab:results}
  \scriptsize
  \setlength{\tabcolsep}{4pt}
  \begin{tabular}{lrrrrrrrr}
    \toprule
    & & & \multicolumn{4}{c}{Model-free} & \multicolumn{2}{c}{Imagination} \\
    \cmidrule(lr){4-7}
    \cmidrule(lr){8-9}
    Game
    & Random
    & Human
    & DER
    & CURL
    & DrQ($\varepsilon$)
    & SPR
    & SimPLe
    & TWM (ours) \\
    \midrule
    Alien & 227.8 & 7127.7 & 802.3 & 711.0 & \textbf{865.2} & 841.9 & 616.9 & 674.6 \\
    Amidar & 5.8 & 1719.5 & 125.9 & 113.7 & 137.8 & \textbf{179.7} & 74.3 & 121.8 \\
    Assault & 222.4 & 742.0 & 561.5 & 500.9 & 579.6 & 565.6 & 527.2 & \textbf{682.6} \\
    Asterix & 210.0 & 8503.3 & 535.4 & 567.2 & 763.6 & 962.5 & \textbf{1128.3} & 1116.6 \\
    BankHeist & 14.2 & 753.1 & 185.5 & 65.3 & 232.9 & 345.4 & 34.2 & \textbf{466.7} \\
    BattleZone & 2360.0 & 37187.5 & 8977.0 & 8997.8 & 10165.3 & \textbf{14834.1} & 4031.2 & 5068.0 \\
    Boxing & 0.1 & 12.1 & -0.3 & 0.9 & 9.0 & 35.7 & 7.8 & \textbf{77.5} \\
    Breakout & 1.7 & 30.5 & 9.2 & 2.6 & 19.8 & 19.6 & 16.4 & \textbf{20.0} \\
    ChopperCommand & 811.0 & 7387.8 & 925.9 & 783.5 & 844.6 & 946.3 & 979.4 & \textbf{1697.4} \\
    CrazyClimber & 10780.5 & 35829.4 & 34508.6 & 9154.4 & 21539.0 & 36700.5 & 62583.6 & \textbf{71820.4} \\
    DemonAttack & 152.1 & 1971.0 & 627.6 & 646.5 & \textbf{1321.5} & 517.6 & 208.1 & 350.2 \\
    Freeway & 0.0 & 29.6 & 20.9 & \textbf{28.3} & 20.3 & 19.3 & 16.7 & 24.3 \\
    Frostbite & 65.2 & 4334.7 & 871.0 & 1226.5 & 1014.2 & 1170.7 & 236.9 & \textbf{1475.6} \\
    Gopher & 257.6 & 2412.5 & 467.0 & 400.9 & 621.6 & 660.6 & 596.8 & \textbf{1674.8} \\
    Hero & 1027.0 & 30826.4 & 6226.0 & 4987.7 & 4167.9 & 5858.6 & 2656.6 & \textbf{7254.0} \\
    Jamesbond & 29.0 & 302.8 & 275.7 & 331.0 & 349.1 & \textbf{366.5} & 100.5 & 362.4 \\
    Kangaroo & 52.0 & 3035.0 & 581.7 & 740.2 & 1088.4 & \textbf{3617.4} & 51.2 & 1240.0 \\
    Krull & 1598.0 & 2665.5 & 3256.9 & 3049.2 & 4402.1 & 3681.6 & 2204.8 & \textbf{6349.2} \\
    KungFuMaster & 258.5 & 22736.3 & 6580.1 & 8155.6 & 11467.4 & 14783.2 & 14862.5 & \textbf{24554.6} \\
    MsPacman & 307.3 & 6951.6 & 1187.4 & 1064.0 & 1218.1 & 1318.4 & 1480.0 & \textbf{1588.4} \\
    Pong & -20.7 & 14.6 & -9.7 & -18.5 & -9.1 & -5.4 & 12.8 & \textbf{18.8} \\
    PrivateEye & 24.9 & 69571.3 & 72.8 & 81.9 & 3.5 & 86.0 & 35.0 & \textbf{86.6} \\
    Qbert & 163.9 & 13455.0 & 1773.5 & 727.0 & 1810.7 & 866.3 & 1288.8 & \textbf{3330.8} \\
    RoadRunner & 11.5 & 7845.0 & 11843.4 & 5006.1 & 11211.4 & \textbf{12213.1} & 5640.6 & 9109.0 \\
    Seaquest & 68.4 & 42054.7 & 304.6 & 315.2 & 352.3 & 558.1 & 683.3 & \textbf{774.4} \\
    UpNDown & 533.4 & 11693.2 & 3075.0 & 2646.4 & 4324.5 & 10859.2 & 3350.3 & \textbf{15981.7} \\
    \midrule
    Normalized Mean & 0.000 & 1.000 & 0.350 & 0.261 & 0.465 & 0.616 & 0.332 & \textbf{0.956} \\
    Normalized Median & 0.000 & 1.000 & 0.189 & 0.092 & 0.313 & 0.396 & 0.134 & \textbf{0.505} \\
    \bottomrule
  \end{tabular}
\end{table}

\subsection{Ablation Studies}

\paragraph{Uniform Sampling:} \label{sec:ablation-uniform-sampling}

\begin{figure}
  \includegraphics[width=\textwidth]{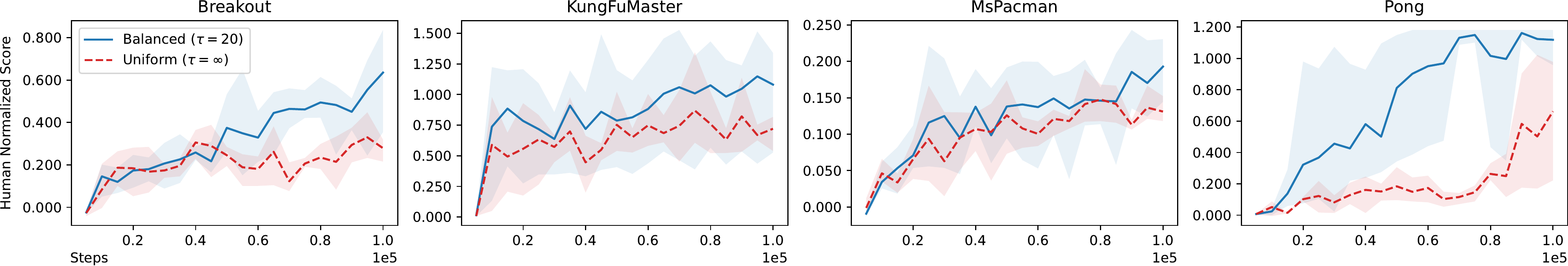}
  \caption{Comparison of the proposed balanced sampling procedure with uniform
  sampling on a random subset of games. We show the development of the human
  normalized score in the course of training. The score is higher with balanced
  sampling, demonstrating its importance.}
  \label{fig:uniform-sampling}
\end{figure}

To show the effectiveness of the sampling procedure described in
\cref{sec:sampling}, we evaluate three games with uniform dataset sampling,
which is equivalent to setting $\datasettemp = \infty$ in \cref{eq:sample-prob}.
In \cref{fig:uniform-sampling} we show that balanced dataset sampling
significantly improves the performance in these games. At the end of training,
the dynamics loss from \cref{eq:dynamics-loss} is lower when applying balanced
sampling. One reason might be that the world model overfits on early training
data and performs bad in later stages of training.

\paragraph{No Rewards:} \label{sec:ablation-no-rewards}

\begin{figure}
  \includegraphics[width=\textwidth]{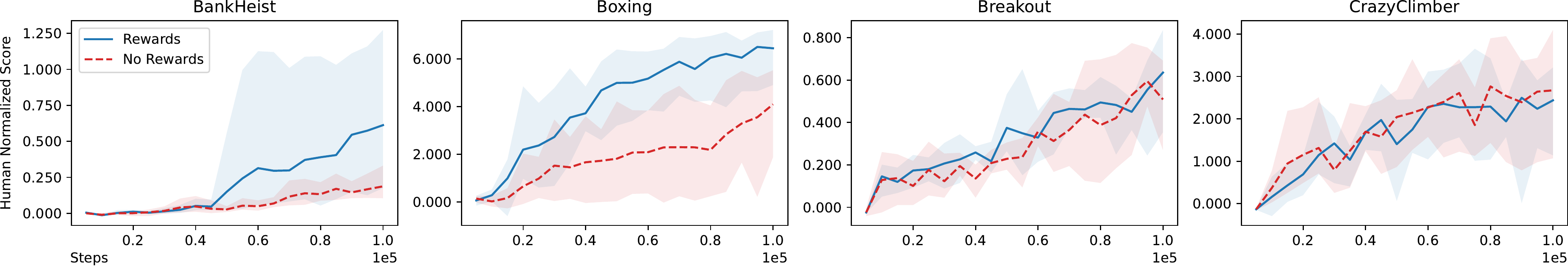}
  \includegraphics[width=\textwidth]{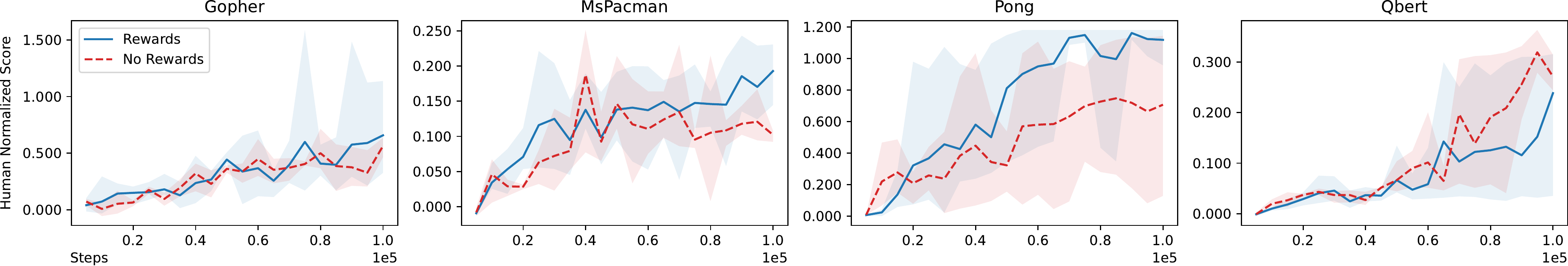}
  \caption{Effect of removing rewards from the input. We show the human
  normalized score during training of a random subset of games. Conditioning on
  rewards can significantly increase the performance. Some games do not benefit
  from the rewards and the score stays roughly the same.}
  \label{fig:no-rewards}
\end{figure}

As described in \cref{sec:dynamics-model}, the predicted rewards are fed back
into the transformer. In \cref{fig:no-rewards} we show on three games that this
can significantly increase the performance. In some games the performance is
equivalent, probably because the world model can make correct predictions solely
based on the latent states and actions.

In \cref{sec:additional-ablation-studies} we perform additional ablation
studies, including the thresholded entropy loss, a shorter history length,
conditioning the policy on $[z,h]$, and increasing the sample efficiency.

\section{Related Work}

The Dyna architecture \citep{dyna} introduced the idea of training a model of
the environment and using it to further improve the value function or the
policy. \citet{world-models} introduced the notion of a \textit{world model},
which tries to completely imitate the environment and is used to generate
experience to train a model-free agent. They implement a world model as a VAE
\citep{vae} and an RNN and learn a policy in latent space with an evolution
strategy. With SimPLe, \citet{simple} propose an iterative training procedure
that alternates between training the world model and the policy. Their policy
operates on pixel-level and is trained using PPO \citep{ppo}. \citet{dreamerv1}
present Dreamer and implement a world model as a stochastic RNN that splits the
latent state in a stochastic part and a deterministic part; this idea was first
introduced by \citealp{planet}. This allows their world model to capture the
stochasticity of the environment and simultaneously facilitates remembering
information over multiple time steps. \cite{robine2020smaller} use a VQ-VAE to
construct a world model with drastically lower number of parameters. DreamerV2
\citep{dreamerv2} achieves great performance on the Atari $50$M benchmark after
making some changes to Dreamer, the most important ones being categorical latent
variables and an improved objective. 

Another direction of model-based reinforcement learning is planning, where the
model is used at inference time to improve the action selection by looking ahead
several time steps into the future. The most prominent work is MuZero
\citep{muzero}, where a learned sequence model of rewards and values is combined
with Monte-Carlo Tree Search \citep{mcts} without learning explicit
representations of the observations. MuZero achieves impressive performance on
the Atari $50$M benchmark, but it is also computationally expensive and requires
significant engineering effort. EfficientZero \citep{efficient-zero} improves
MuZero and achieves great performance on the Atari 100k benchmark.

Transformers \citep{transformer} advanced the effectiveness of sequence models
in multiple domains, such as natural language processing and computer vision
\citep{vit}. Recently, they have also been applied to reinforcement learning
tasks. The Decision Transformer \citep{decision-transformer} and the Trajectory
Transformer \citep{trajectory-transformer} are trained on an offline dataset of
trajectories. The Decision Transformer is conditioned on states, actions, and
returns, and outputs optimal actions. The Trajectory Transformer trains a
sequence model of states, actions, and rewards, and is used for planning.
\cite{transdreamer} replace the RNN of Dreamer with a transformer and outperform
Dreamer on Hidden Order Discovery tasks. However, their transformer has no
access to previous rewards and they do not evaluate their method on the Atari
100k benchmark. Moreover, their policy depends on the outputs of the
transformer, leading to higher computational costs during inference time.
Concurrent to and independent from our work, \citet{iris} apply a transformer to
sequences of frame tokens and actions and achieve state-of-the-art results on
the Atari 100k benchmark.

\section{Conclusion}
In this work, we discuss a reinforcement learning approach using
transformer-based world models. Our method (TWM) outperforms previous model-free
and model-based methods in terms of human normalized score on the 26 games of
the Atari 100k benchmark. By using the transformer only during training, we were
able to keep the computational costs low during inference, i.e., when running
the learned policy in a real environment. We show how feeding back the predicted
rewards into the transformer is beneficial for learning the world model.
Furthermore, we introduce the balanced cross-entropy loss for finer control over
the trade-off between the entropy and cross-entropy terms in the loss functions
of the world model. A new thresholded entropy loss effectively stabilizes the
entropy of the policy. Finally, our novel balanced sampling procedure corrects
issues of naive uniform sampling of past experience.

\clearpage

\bibliography{main}
\bibliographystyle{iclr2023_conference}

\clearpage
\appendix
\section{Appendix}

\subsection{Extended Experiments} \label{sec:extended-experiments}

\begin{figure}[H]
  \captionsetup[subfigure]{aboveskip=3pt,belowskip=0pt}

  \begin{subfigure}{\textwidth}
    \includegraphics[width=\textwidth,trim=5189 60 3462 18,clip]{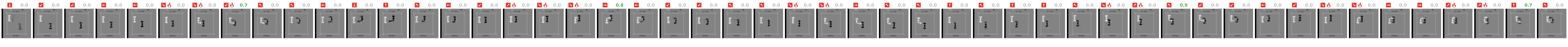}
    \caption{Boxing. The player (white) presses \textit{fire}, misses the
    opponent, gets no reward, and retreats.}
    \vspace{0.1cm}
  \end{subfigure}
  \begin{subfigure}{\textwidth}
    \includegraphics[width=\textwidth,trim=6268 60 2383 18,clip]{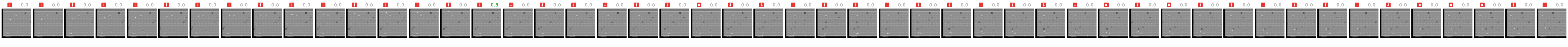}
    \caption{Freeway. The player moves up, steps back because of an approaching
    car, and continues to move up.}
    \vspace{0.1cm}
  \end{subfigure}
  \begin{subfigure}{\textwidth}
    \includegraphics[width=\textwidth,trim=3679 60 4974 18,clip]{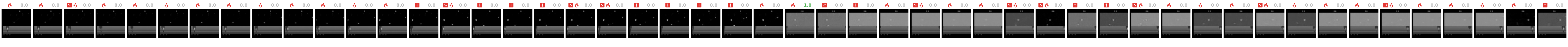}
    \caption{Jamesbond. The player presses \textit{fire} to fire a missile, and
    gets a reward when it hits.}
    \vspace{0.1cm}
  \end{subfigure}
  \begin{subfigure}{\textwidth}
    \includegraphics[width=\textwidth,trim=3896 60 4756 18,clip]{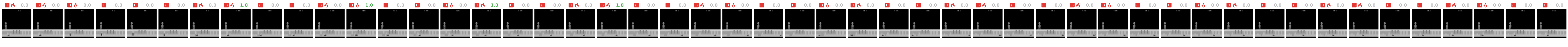}
    \caption{Gopher. The player (farmer) closes a hole created by the opponent
    (gopher). The gopher keeps moving independent from the selected actions,
    indicating that the world model has correctly learned the correlations
    between the player and the actions. Note that the gopher briefly disappears
    and reappears on the other side.}
  \end{subfigure}

  \caption{Additional example trajectories generated by our world model.}
  \label{fig:more-trajectories}
\end{figure}

\vspace{-0.2cm}

\begin{figure}
  \captionsetup[subfigure]{aboveskip=3pt,belowskip=0pt,width=0.75\textwidth}

  \begin{subfigure}{\textwidth}
    \centering
    \includegraphics[width=0.75\textwidth,trim=8 8 6 8]{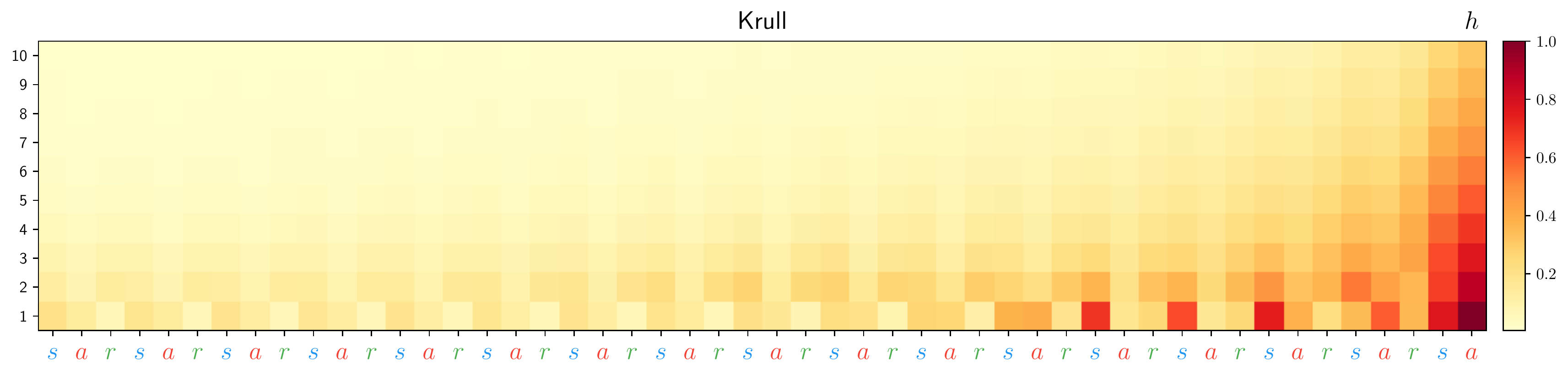}
    \caption{This world model focuses on previous states.}
    \vspace{0.1cm}
  \end{subfigure}
  \begin{subfigure}{\textwidth}
    \centering
    \includegraphics[width=0.75\textwidth,trim=8 8 6 8]{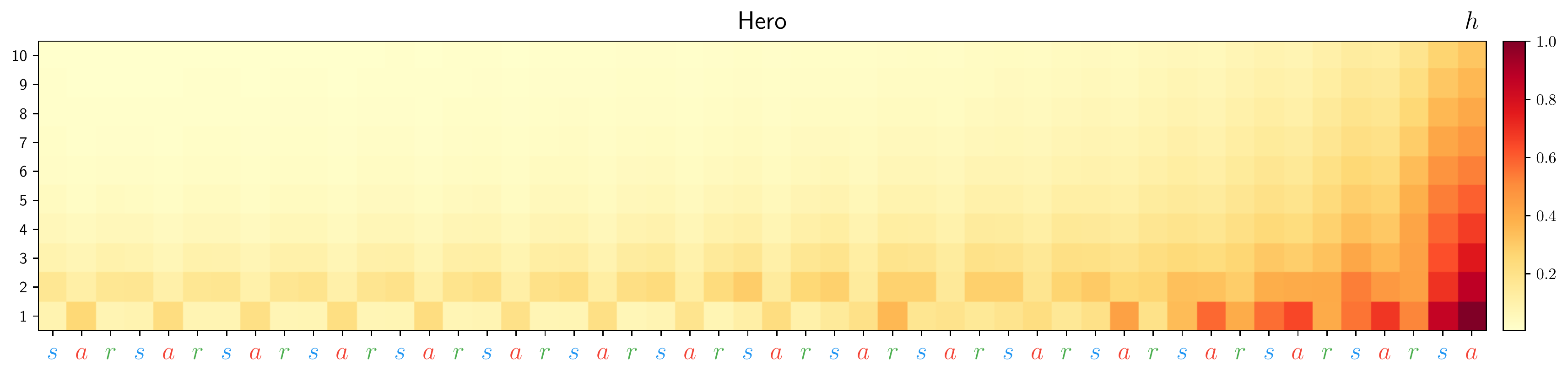}
    \caption{This world model focuses on previous actions, indicating that the
      effect of actions can last longer than a single time step.}
    \vspace{0.1cm}
  \end{subfigure}
  \begin{subfigure}{\textwidth}
    \centering
    \includegraphics[width=0.75\textwidth,trim=8 8 6 8]{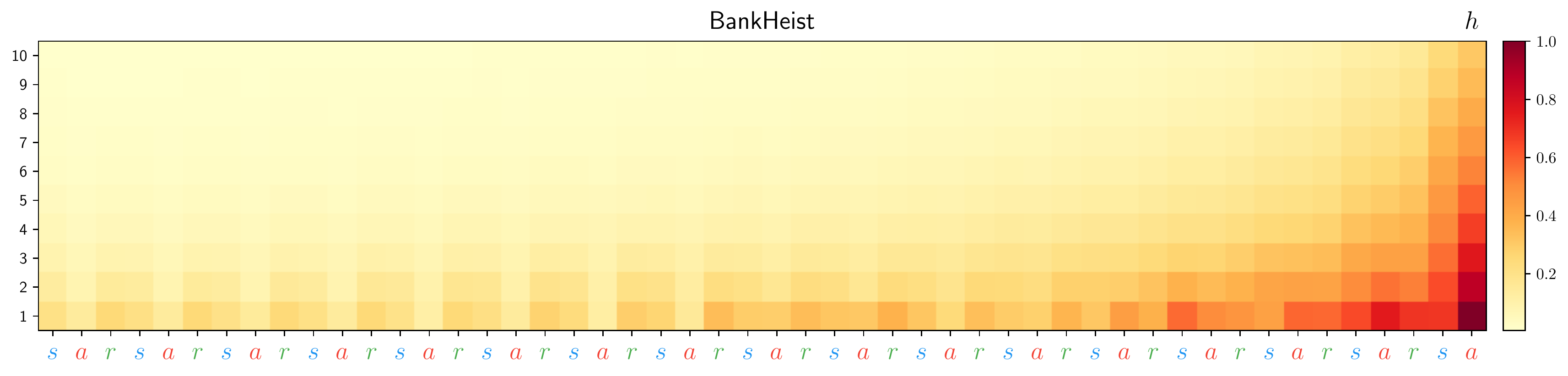}
    \caption{This world model attends to all three modalities in the recent past.}
    \vspace{0.1cm}
  \end{subfigure}
  \begin{subfigure}{\textwidth}
    \centering
    \includegraphics[width=0.75\textwidth,trim=8 8 6 8]{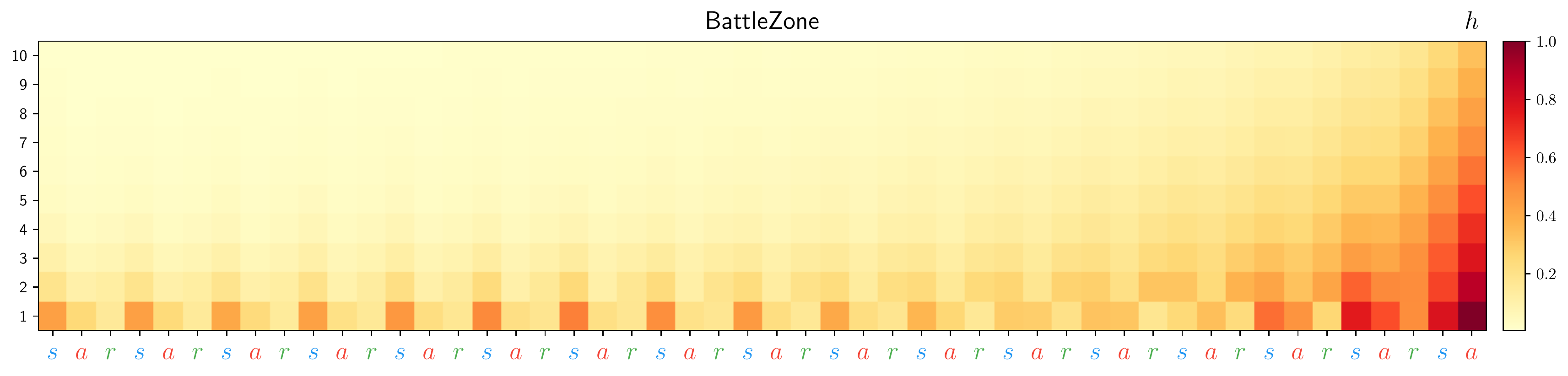}
    \caption{This world model attends to states at all time steps, probably
      because of the complexity of this 3D game.}
    \vspace{0.1cm}
  \end{subfigure}
  \begin{subfigure}{\textwidth}
    \centering
    \includegraphics[width=0.75\textwidth,trim=8 8 6 8]{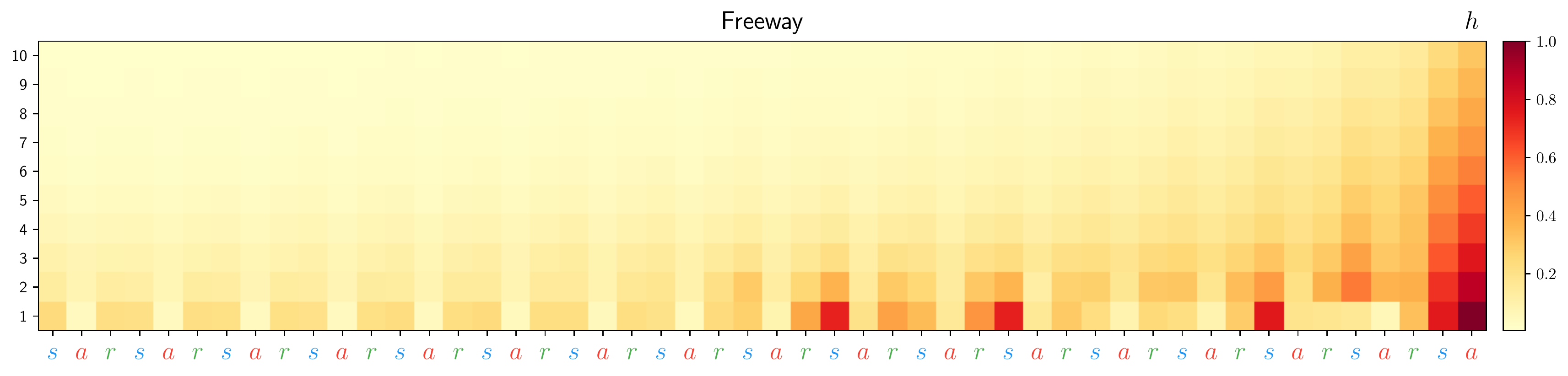}
    \caption{This world model mainly focuses on four states at specific time
      steps.}
    \label{fig:average-attention-map-freeway}
  \end{subfigure}

  \caption{Average attention maps of the transformer, computed over many time steps.
    They show how different games require a different focus on modalities and
    time steps.}
  \label{fig:average-attention-maps}
\end{figure}

\begin{figure}
  \centering
  \includegraphics[width=0.75\textwidth,trim=8 8 6 8]{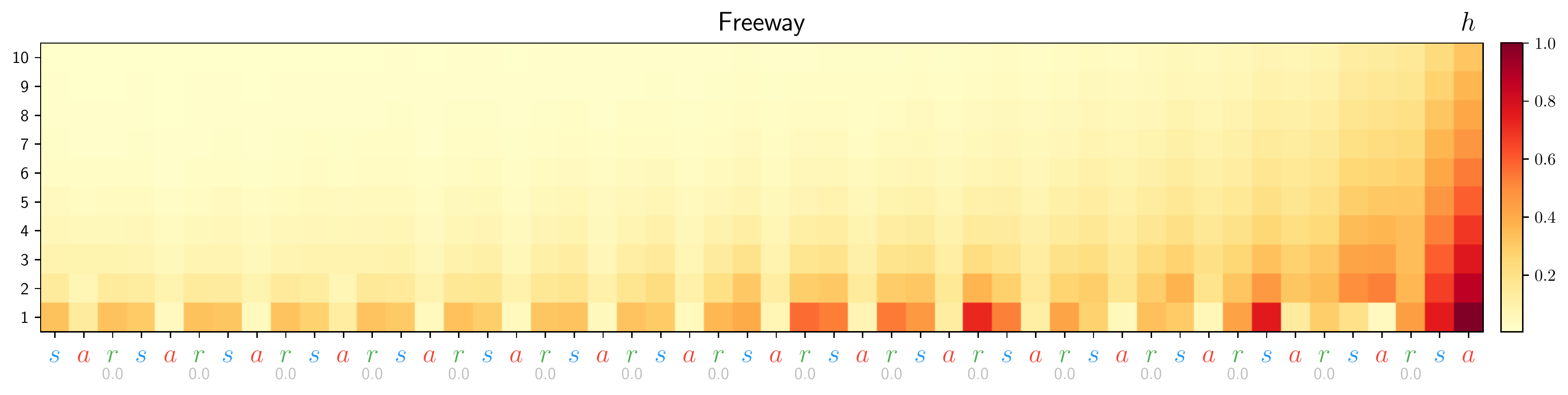}
  \caption{Attention map for Freeway for a single time step. At this point the
    player hits a car and gets pushed back (see also
    \cref{fig:trajectory-freeway}) and the world model puts more attention to
    past states and rewards, compared with the average attention at other time
    steps, as shown in \cref{fig:average-attention-map-freeway}. The world model
    has learned to handle this situation separately.}
  \label{fig:more-attention-maps}
\end{figure}

\begin{figure}
  \includegraphics[width=\textwidth,trim=10 60 10 18,clip]{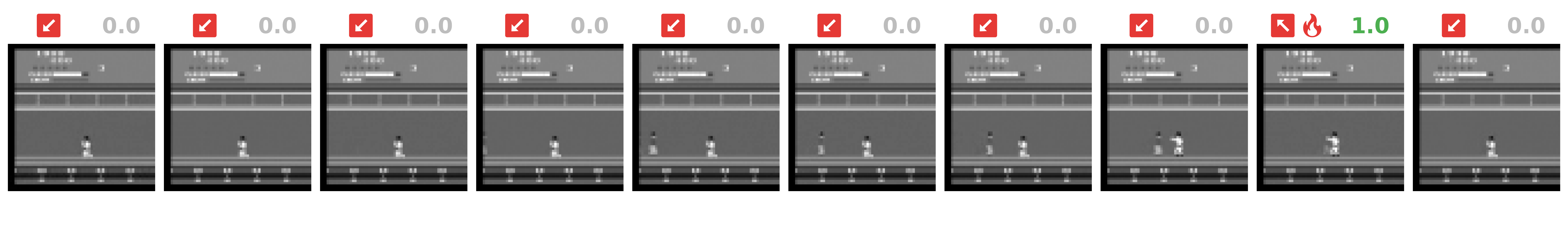}

  \vspace{0.075cm}
  \hrule
  \vspace{0.05cm}

  \includegraphics[width=\textwidth,trim=10 60 10 18,clip]{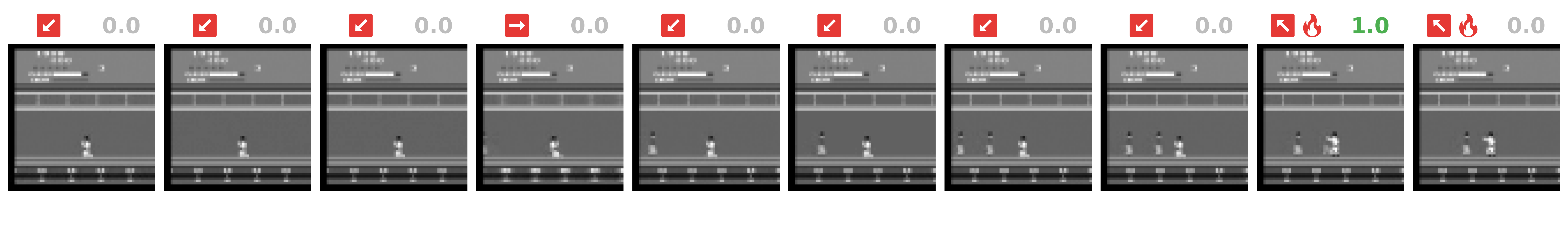}

  \vspace{0.075cm}
  \hrule
  \vspace{0.05cm}

  \includegraphics[width=\textwidth,trim=10 60 10 18,clip]{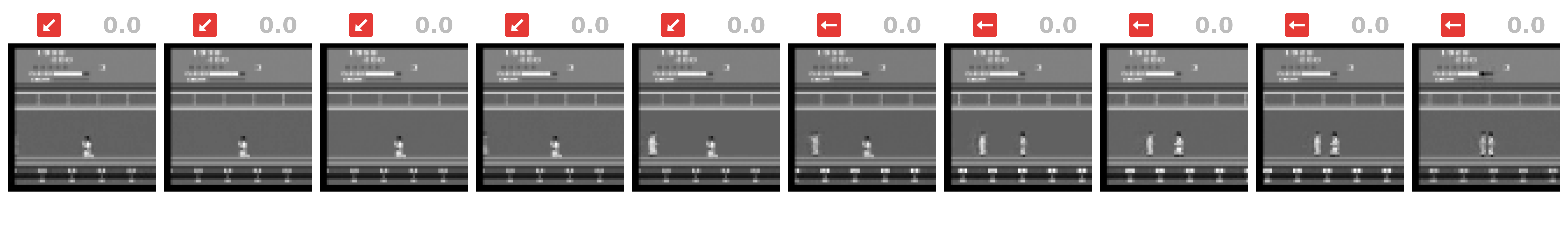}

  \caption{Three trajectories for the game KungFuMaster generated by our world
  model, using the same starting state. Because of its stochastic nature, the
  world model is able to generate three different situations (one opponent, two
  opponents, one other type of opponent). Note that we only show every third
  frame to cover more time steps.}
  \label{fig:stochasticity}
\end{figure}

\begin{figure}
  \includegraphics[width=\textwidth,trim=12 32 11 13,clip]{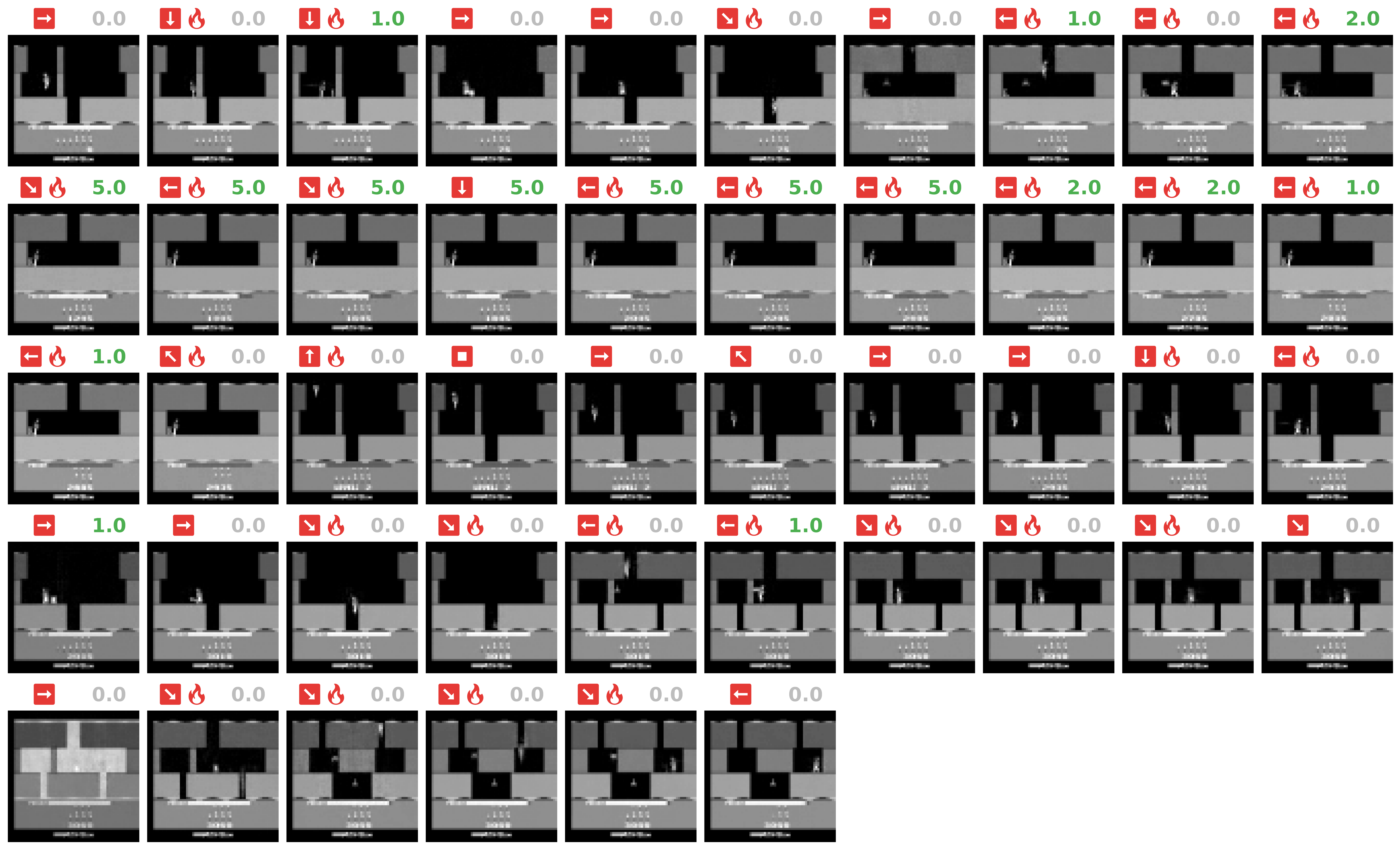}
  \caption{A long trajectory imagined by our world model for the game Hero. The
  player traverses five different rooms and the world model is able to correctly
  predict the state and reward dynamics. Note that we only show every fifth
  frame to cover more time steps (the rewards lying in-between are summed up).
  The total number of time steps is $230$.}
  \label{fig:long-sequence}
\end{figure}

\begin{figure}
  \centering
  \begin{subfigure}{0.17\textwidth}
    \includegraphics[width=\linewidth]{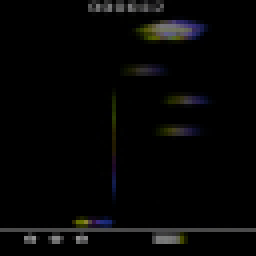}
    \caption{Assault}
  \end{subfigure}
  \hspace{0.1cm}
  \begin{subfigure}{0.17\textwidth}
    \includegraphics[width=\linewidth]{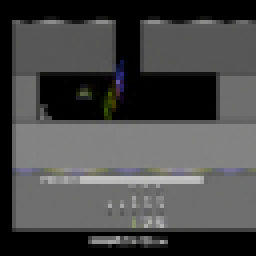}
    \caption{Hero}
  \end{subfigure}
  \hspace{0.1cm}
  \begin{subfigure}{0.17\textwidth}
    \includegraphics[width=\linewidth]{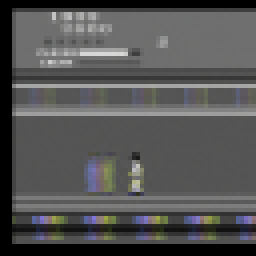}
    \caption{KungFuMaster}
  \end{subfigure}
  \hspace{0.1cm}
  \begin{subfigure}{0.17\textwidth}
    \includegraphics[width=\linewidth]{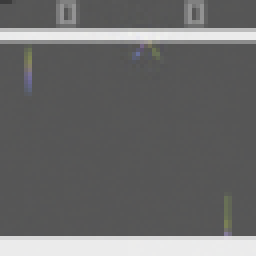}
    \caption{Pong}
  \end{subfigure}
  \hspace{0.1cm}
  \begin{subfigure}{0.17\textwidth}
    \includegraphics[width=\linewidth]{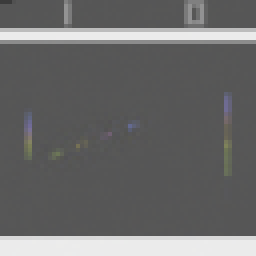}
    \caption{Pong}
  \end{subfigure}
  \caption{Visualization of frame stacks reconstructed from predicted states
    $\hat{z}_t$. Each frame in the stack is visualized by a different color. The
    world model is able to encode and predict movements.}
  \label{fig:frame-stacking}
\end{figure}

\paragraph{Additional Analysis:} \label{sec:additional-analysis}
\begin{enumerate}
  \vspace{-0.5\baselineskip}
  \item We provide more example trajectories in \cref{fig:more-trajectories}.
  \item We present more attention plots in
    \cref{fig:average-attention-maps,fig:more-attention-maps}. All attention
    maps are generated using the attention rollout method by
    \citet{attention-flow}. Note that we had to modify the method slightly, in
    order to take the causal masks into account.
  \item \textit{Sample Efficiency}: We provide the scores of our main
    experiments after different amounts of interactions with the environment in
    \cref{tab:sample-efficiency}. After $50$K interactions, our method already
    has a higher mean normalized score than previous sample-efficient methods.
    Our mean normalized score is higher than DER, CURL, and SimPLe after $25$K
    interactions. This demonstrates the high sample efficiency of our approach.
  \item \textit{Stochasticity}: The stochastic prediction of the next state
    allows the world model to sample a variety of trajectories, even from the
    same starting state, as can be seen in \cref{fig:stochasticity}.
  \item \textit{Long Sequence Imagination}: The world model is trained using
    sequences of length $\historylength = 16$, however, it generalizes well to
    very long trajectories, as shown in \cref{fig:long-sequence}.
  \item \textit{Frame Stacking}: In \cref{fig:frame-stacking} we visualize the
    learned stacks of frames. This shows that the world model encodes and
    predicts the motion of objects.
\end{enumerate}

\begin{figure}
  \begin{subfigure}{\textwidth}
    \centering
    \includegraphics[width=\textwidth]{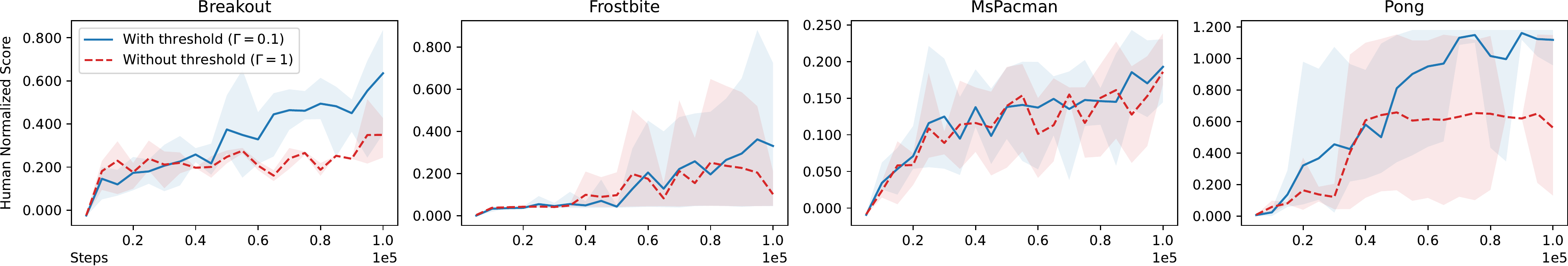}
    \vspace{-0.24cm}
  \end{subfigure}
  \begin{subfigure}{\textwidth}
    \centering
    \includegraphics[width=0.55\textwidth]{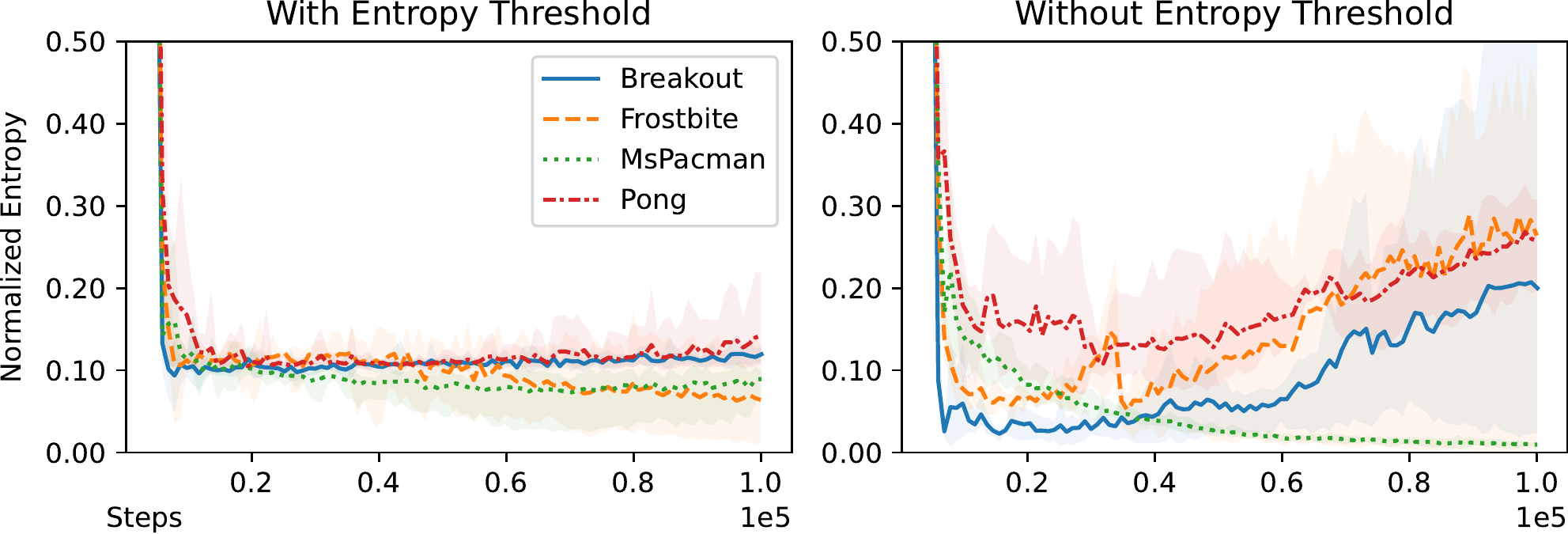}
  \end{subfigure}

  \caption{Effect of disabling the proposed thresholded entropy loss (by setting
    $\actorentropythreshold = 1$) on the performance and the entropy in a random
    subset of games. The thresholded version stabilizes the entropy and leads to
    a better score in Breakout and Pong, while the entropy behaves unfavorably
    without a threshold.}
  \label{fig:entropy-threshold}
\end{figure}

\begin{figure}
  \includegraphics[width=\textwidth]{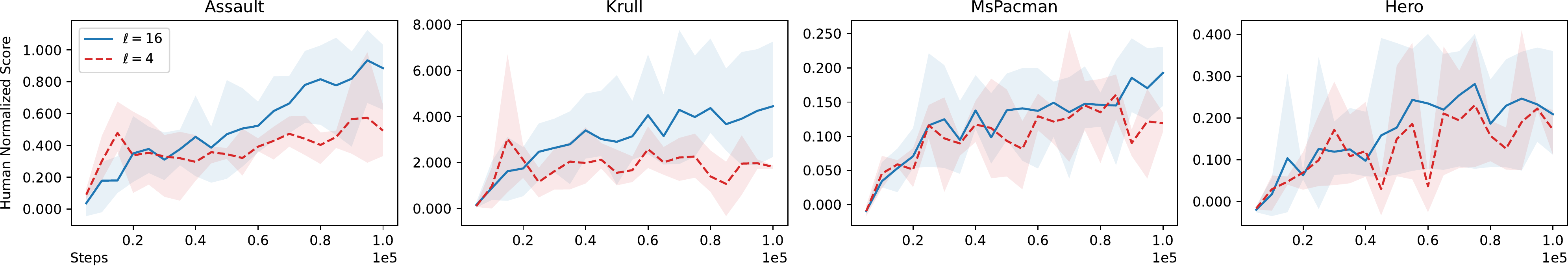}
  \caption{Comparison of the history length $\historylength = 16$ used in our
    main experiments with $\historylength = 4$ on a random subset of games. We
    observe a lower human normalized score for $\historylength = 4$.}
  \label{fig:history-length}
\end{figure}

\begin{figure}
  \includegraphics[width=\textwidth]{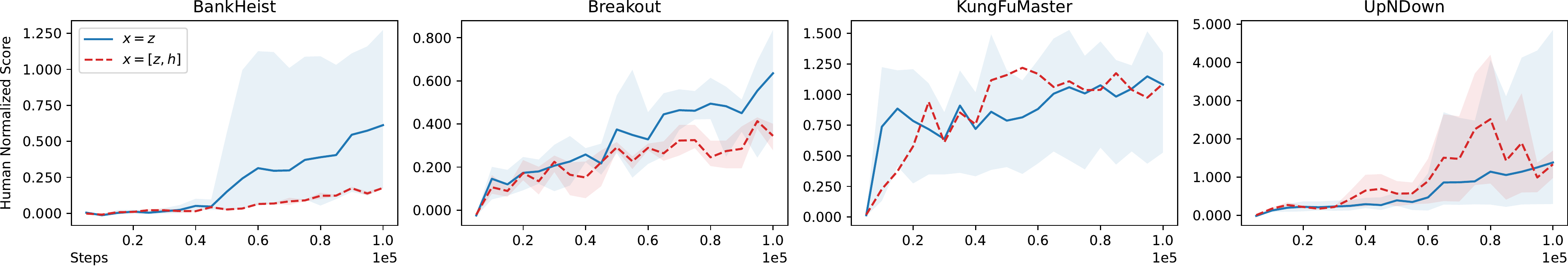}
  \caption{Conditioning the policy on $[z,h]$ compared with the usual $z$. In
    some cases the performance can be better during training, but the final
    score is lower or equal.}
  \label{fig:hidden-input}
\end{figure}

\begin{figure}
  \centering
  \includegraphics[width=\textwidth]{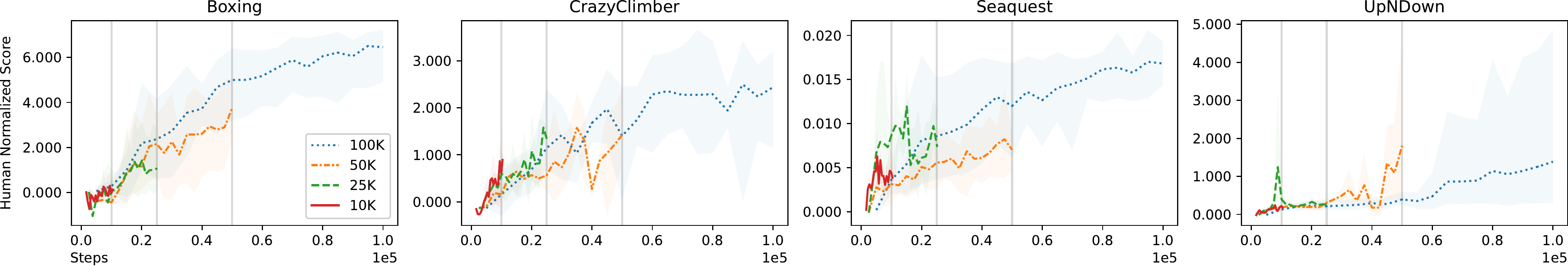}
  \caption{Scores on a random subset of games when we train with a lower number
    of interactions but the same training budget. This only leads to a
    significant improvement for UpNDown, where the final score is higher with
    only $50$K interactions.}
  \label{fig:sample-efficiency}
\end{figure}

\paragraph{Additional Ablation Studies:} \label{sec:additional-ablation-studies}
\begin{enumerate}
  \vspace{-0.5\baselineskip}
  \item \textit{Thresholded Entropy Loss}: In \cref{fig:entropy-threshold} we
    compare (i) our thresholded entropy loss for the policy (see
    \cref{sec:entropy-threshold}) with (ii) the usual entropy penalty. For (i)
    we use the same hyperparameters as in our main experiments, i.e.,
    $\actorentropycoef = 0.01$ and $\actorentropythreshold = 0.1$. For (ii) we
    set $\actorentropycoef = 0.001$ and $\actorentropythreshold = 1.0$, which
    effectively disables the threshold. Without a threshold, the entropy is more
    likely to either collapse or diverge. When the threshold is used, the score
    is higher as well, probably because the entropy is in a more sensible range
    for the exploration-exploitation trade-off. This cannot be solved by
    adjusting the penalty coefficient $\actorentropycoef$ alone, since it would
    increase or decrease the entropy in all games.
  \item \textit{History Length}: We trained our world model with a shorter
    history and set $\historylength = 4$ instead of $\historylength = 16$. This
    has a negative impact on the score, as can be seen in
    \cref{fig:history-length}, demonstrating that more time steps into the past
    are important.
  \item \textit{Choice of Policy Input}: In \cref{sec:policy-input} we explained
    why the input to the policy is only the latent state, i.e., $x = z$. In
    \cref{fig:hidden-input} we show that using $x = [z, h]$ can result in lower
    final scores. We hypothesize that the policy network has a hard time keeping
    up with the changes of the space of $h$ during training and cannot ignore
    this additional information.
  \item \textit{Increasing the Sample Efficiency}: To find out whether we can
    further increase the sample efficiency shown in
    \cref{tab:sample-efficiency}, we train a random subset of games again on
    $10$K, $25$K, and $50$K interactions with the full training budget that we
    used for the $100$K interactions. In \cref{fig:sample-efficiency} we see
    that this can lead to significant improvements in some cases, which could
    mean that the policy benefits from more training on imagined trajectories,
    but can even lead to worse performance in other cases, which could possibly
    be caused by overfitting of the world model. When the performance stays the
    same even with longer training, this could mean that better exploration in
    the real environment is required to get further improvements.
\end{enumerate}

\begin{table}[H]
  \centering
  \caption{Performance of our method at different stages of training compared
  with final scores of previous methods. We show individual game scores and mean
  human normalized scores. The normalized mean of our method is higher than
  SimPLe after only $25$K interactions, and higher than previous methods after
  $50$K interactions.}
  \label{tab:sample-efficiency}
  \scriptsize
  \setlength{\tabcolsep}{4pt}
  \begin{tabular}{lrrrrrrrrrr}
    \toprule
    & & & & & \multicolumn{6}{c}{TWM (ours)} \\
    \cmidrule(lr){6-11}
    Game           &  Random &   Human &   SimPLe &      SPR &   $5$K &   $10$K &   $25$K &   $50$K &   $75$K &  $100$K \\
    \midrule
    Alien          &   227.8 &  7127.7 &    616.9 &    841.9 &  202.8 &   383.2 &   463.6 &   532.0 &   776.6 &   674.6 \\
    Amidar         &     5.8 &  1719.5 &     74.3 &    179.7 &    3.8 &    35.4 &    54.9 &   101.3 &   103.0 &   121.8 \\
    Assault        &   222.4 &   742.0 &    527.2 &    565.6 &  241.5 &   315.4 &   418.7 &   466.8 &   627.8 &   682.6 \\
    Asterix        &   210.0 &  8503.3 &   1128.3 &    962.5 &  277.0 &   297.0 &   536.0 &   912.0 &   886.0 &  1116.6 \\
    BankHeist      &    14.2 &   753.1 &     34.2 &    345.4 &   17.6 &     4.4 &    17.4 &   125.2 &   288.4 &   466.7 \\
    BattleZone     &  2360.0 & 37187.5 &   4031.2 &  14834.1 & 2640.0 &  3120.0 &  2700.0 &  3740.0 &  5260.0 &  5068.0 \\
    Boxing         &     0.1 &    12.1 &      7.8 &     35.7 &    0.8 &     3.4 &    28.5 &    60.1 &    67.1 &    77.5 \\
    Breakout       &     1.7 &    30.5 &     16.4 &     19.6 &    1.0 &     5.9 &     6.9 &    12.5 &    15.0 &    20.0 \\
    ChopperCommand &   811.0 &  7387.8 &    979.4 &    946.3 &  928.0 &  1044.0 &  1358.0 &  1306.0 &  1438.0 &  1697.4 \\
    CrazyClimber   & 10780.5 & 35829.4 &  62583.6 &  36700.5 & 7425.0 & 14773.2 & 39456.8 & 45916.0 & 67766.2 & 71820.4 \\
    DemonAttack    &   152.1 &  1971.0 &    208.1 &    517.6 &  174.7 &   184.4 &   216.8 &   335.2 &   391.4 &   350.2 \\
    Freeway        &     0.0 &    29.6 &     16.7 &     19.3 &    0.0 &     4.6 &    20.8 &    23.7 &    23.9 &    24.3 \\
    Frostbite      &    65.2 &  4334.7 &    236.9 &   1170.7 &   66.2 &   204.6 &   297.8 &   247.6 &  1165.4 &  1475.6 \\
    Gopher         &   257.6 &  2412.5 &    596.8 &    660.6 &  345.2 &   414.0 &   593.2 &  1213.2 &  1549.2 &  1674.8 \\
    Hero           &  1027.0 & 30826.4 &   2656.6 &   5858.6 &  448.9 &  1552.6 &  4790.9 &  6302.7 &  9403.8 &  7254.0 \\
    Jamesbond      &    29.0 &   302.8 &    100.5 &    366.5 &   35.0 &   117.0 &   172.0 &   215.0 &   322.0 &   362.4 \\
    Kangaroo       &    52.0 &  3035.0 &     51.2 &   3617.4 &   28.0 &    92.0 &   476.0 &   724.0 &   876.0 &  1240.0 \\
    Krull          &  1598.0 &  2665.5 &   2204.8 &   3681.6 & 1763.6 &  2552.8 &  4234.0 &  4699.2 &  5848.0 &  6349.2 \\
    KungFuMaster   &   258.5 & 22736.3 &  14862.5 &  14783.2 &  574.0 & 16828.0 & 16368.0 & 17946.0 & 22936.0 & 24554.6 \\
    MsPacman       &   307.3 &  6951.6 &   1480.0 &   1318.4 &  245.9 &   535.1 &  1077.5 &  1224.3 &  1287.6 &  1588.4 \\
    Pong           &   -20.7 &    14.6 &     12.8 &     -5.4 &  -20.4 &   -19.8 &    -7.7 &     8.0 &    19.9 &    18.8 \\
    PrivateEye     &    24.9 & 69571.3 &     35.0 &     86.0 &   61.0 &    80.0 &    80.0 &     3.2 &    88.8 &    86.6 \\
    Qbert          &   163.9 & 13455.0 &   1288.8 &    866.3 &  151.0 &   298.5 &   703.5 &  1046.5 &  1788.5 &  3330.8 \\
    RoadRunner     &    11.5 &  7845.0 &   5640.6 &  12213.1 &   24.0 &  1120.0 &  5178.0 &  7436.0 &  8034.0 &  9109.0 \\
    Seaquest       &    68.4 & 42054.7 &    683.3 &    558.1 &   76.8 &   221.2 &   428.4 &   572.0 &   704.0 &   774.4 \\
    UpNDown        &   533.4 & 11693.2 &   3350.3 &  10859.2 &  385.8 &  1963.0 &  2905.6 &  4922.8 & 10478.6 & 15981.7 \\
    \midrule
    Normalized Mean & 0.000 & 1.000 & 0.332 & 0.616 & 0.007 & 0.133 & 0.408 & 0.624 & 0.832 & 0.956 \\
    \bottomrule
  \end{tabular}
  \vspace{-0.2cm}
\end{table}

\paragraph{Wall-Clock Times:}
For each run, we give the agent a total training and evaluation budget of
roughly $10$ hours on a single NVIDIA A100 GPU. The time can vary slightly,
since the budget is based on the number of updates. An NVIDIA GeForce RTX 3090
requires 12-13 hours for the same amount of training and evaluation. When using
a vanilla transformer, which does not use the memory mechanism of the
Transformer-XL architecture \citep{transformer-xl}, the runtime is roughly
$15.5$ hours on an NVIDIA A100 GPU, i.e., $1.5$ times higher.

We compare the runtime of our method with previous methods in
\cref{tab:wall-clock-times}. Our method is more than $20$ times faster than
SimPLe, but slower than model-free methods. However, our method should be as
fast as other model-free methods during inference. In
\cref{tab:sample-efficiency} we have shown that our method achieves a higher
human normalized score than previous sample-efficient methods after 50K
interactions. This suggests that our method could potentially outperform
previous methods with shorter training, which would take less than $23.3$ hours.

To determine how time-consuming the individual parts of our method are, we
investigate the throughput of the models, with the batch sizes of our main
experiments. The Transformer-XL version is almost twice as fast, which again
shows the importance of this design choice. The throughputs were measured on an
NVIDIA A100 GPU and are given in (approximate) samples per second:
\begin{itemize}
  \item World model training: 16,800 samples/s
  \item World model imagination (Transformer-XL): 39,000 samples/s
  \item World model imagination (vanilla): 19,900 samples/s
  \item Policy training: 700,000 samples/s
\end{itemize}
We also examine how fast the policy can run in an Atari game. We measured the
(approximate) frames per second on a CPU (since the batch size is $1$).
Conditioning the policy on $[z, h]$ is about $3$ times slower than $z$, since
the transformer is required:
\begin{itemize}
  \item Policy conditioned on $z$: 653 frames/s
  \item Policy conditioned on $[z,h]$: 213 frames/s
\end{itemize}

\begin{table}
  \caption{Approximate runtime (i.e., training and evaluation time for a single
    run) of our method compared with previous methods that also evaluate on the
    Atari 100k benchmark. Runtimes of previous methods are taken from
    \citet{spr}. They used an improved version of DER
    \citep{data-efficient-rainbow}, which is roughly equivalent to DrQ
    \citep{drq}, so the specified runtime might differ from the original DER
    implementation. There are data augmented versions for SPR and DER. All
    runtimes are measured on a single NVIDIA P100 GPU.}
  \label{tab:wall-clock-times}
  \centering
  \footnotesize
  \begin{tabular}{lcr}
    \toprule
    Method & Model-based & Runtime in hours \\
    \midrule
    SimPLe & \ding{51} & 500 \\
    TWM (ours) & \ding{51} & 23.3 \\
    SPR (with aug.) & \ding{55} & 4.6 \\
    SPR (w/o aug.) & \ding{55} & 3.0 \\
    DER/DrQ (with aug.) & \ding{55} & 2.1 \\
    DER/DrQ (w/o aug.) & \ding{55} & 1.4 \\
    \bottomrule
  \end{tabular}
  \vspace{-0.3cm}
\end{table}

\begin{figure}
  \includegraphics[width=\textwidth]{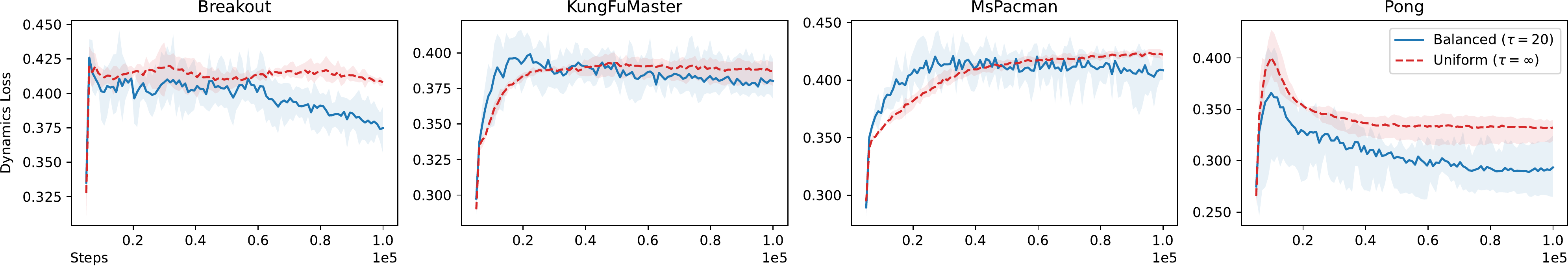}
  \caption{Comparison of the proposed balanced sampling procedure with uniform
  sampling. It shows the development of the dynamics loss from
  \cref{eq:dynamics-loss}, which is lower at the end of training in all cases.}
\end{figure}

\subsection{Derivation of Balanced Cross-Entropy Loss} \label{sec:loss-derivation}

\citet{dreamerv2} propose to use a balanced KL divergence loss to jointly
optimize the observation encoder $q_\theta$ and state predictor $p_\theta$ with
shared parameters $\theta$, i.e.,
\begin{equation}
  \lambda \, \kl{\sg{q_\theta}}{p_\theta} + (1 - \lambda) \, \kl{q_\theta}{\sg{p_\theta}},
\end{equation}
where $\sg{\cdot}$ denotes the stop-gradient operation and $\lambda \in [0,1]$
controls how much the state predictor adapts to the observation encoder and vice
versa. We use the identity $\kl{q}{p} = \ent{q,p} - \ent{q}$, where $\ent{q,p}$
is the cross-entropy of distribution $p$ relative to distribution $q$, and show
that our loss functions lead to the same gradients as the balanced KL objective,
but with finer control over the individual components:
\begin{align}
  & \nabla_\theta \left[ \lambda \, \kl{\sg{q_\theta}}{p_\theta} + (1 - \lambda) \, \kl{q_\theta}{\sg{p_\theta}} \right] \\
  ={} & \nabla_\theta \left[ \lambda \left(\ent{\sg{q_\theta},p_\theta} - \ent{\sg{q_\theta}} \right) + (1 - \lambda) \left( \ent{q_\theta,\sg{p_\theta}} - \ent{q_\theta} \right) \right] \\
  ={} & \nabla_\theta \left[ \lambda_1 \, \ent{\sg{q_\theta},p_\theta} + \lambda_2 \, \ent{q_\theta,\sg{p_\theta}} - \lambda_3 \, \ent{q_\theta} \right]\!, \label{eq:combined-gradient}
\end{align}
since $\nabla_\theta \ent{\sg{q_\theta}} = 0$ and by defining $\lambda_1 =
\lambda$ and $\lambda_2 = 1 - \lambda$ and $\lambda_3 = 1 - \lambda$. In this
form, we have control over the cross-entropy of the state predictor relative to
the observation encoder and vice versa. Moreover, we explicitly penalize the
entropy of the observation encoder, instead of being entangled inside of the KL
divergence.

As common in the literature, we define the loss function by omitting the
gradient in \cref{eq:combined-gradient}, so that automatic differentiation
computes this gradient. For our world model, we split the objective into two
loss functions, as the observation encoder and state predictor have separate
parameters, yielding \cref{eq:observation-loss,eq:dynamics-loss}.

\subsection{Additional Training Details}

In \cref{algo:main} we present pseudocode for training the world model and the
actor-critic agent. We use the SiLU activation function \citep{silu} for all
models. In \cref{tab:hyperparameters} we summarize all hyperparameters that we
used in our experiments. In \cref{tab:model-size} we provide the number of
parameters of our models.

\paragraph{Pretraining for Better Initialization:}
During training we need to correctly balance the amount of world model training
and policy training, since the policy has to keep up with the distributional
shift of the latent space. However, we can spend some extra training time on the
world model with pre-collected data (included in the $100$K interactions) at the
beginning of training in order to obtain a reasonable initialization for the
latent states.

\begin{algorithm}
  \caption{Training the world model and the actor-critic agent.}
  \label{algo:main}
  \vspace{0.3em}
  \fontfamily{cmtt}\selectfont
  \footnotesize
  \begin{minipage}[t]{0.49\textwidth}
  \begin{algorithmic}
    \Function{train\_world\_model}{\,}
      \LineComment{sample sequences of observations,}
      \LineComment{rewards, actions and discounts}
      \State o,a,r,d = sample\_from\_dataset()
      \State z = encode(o)
      \State o\_hat = decode(z)
      \State h = transformer(z,a,r)
      \State r\_hat,d\_hat,z\_hat = predict(h)
      \vspace{0.5em}
      \LineComment{optimize world model via}
      \LineComment{self-supervised learning}
      \State optim\_observation(o,z,o\_hat,z\_hat)
      \State optim\_dynamics(r,d,z,r\_hat,d\_hat,z\_hat)
      \vspace{-0.5em}
      \LineComment{z will be used for imagination}
      \State \Return z
    \EndFunction
  \end{algorithmic}
  \end{minipage}
  \vspace{0.5em}
  \hfill
  \begin{minipage}[t]{0.49\textwidth}
    \begin{algorithmic}
      \Function{train\_actor\_critic}{z}
        \LineComment{imagine trajectories of states,}
        \LineComment{rewards, actions and discounts;}
        \LineComment{use z as starting point}
        \State imag = [z]
        \For{t = 0 \keywordfont{until} $\achorizon$}
          \State a = actor(z)
          \State imag.append(a)
          \State h = transformer(imag)
          \State r,d,z = predict(h)
          \State imag.extend([r,d,z])
        \EndFor
        \vspace{0.5em}
        \LineComment{optimize actor-critic via}
        \LineComment{reinforcement learning}
        \State optim\_actor\_critic(imag)
      \EndFunction
    \end{algorithmic}
  \end{minipage}
\end{algorithm}

\begin{table}
  \centering
  \caption{Hyperparameters used in our experiments.}
  \label{tab:hyperparameters}
  \footnotesize
  \begin{tabular}{lcr}
    \toprule
    Description & Symbol & Value \\
    \midrule
    Dataset sampling temperature & $\datasettemp$ & 20 \\
    Discount factor & $\gamma$ & 0.99 \\
    GAE parameter & $\lambda$ & 0.95 \\
    World model batch size & $\wmbatchsize$ & 100 \\
    History length & $\historylength$ & 16 \\
    Imagination batch size & $\acbatchsize$ & 400 \\
    Imagination horizon & $\achorizon$ & 15 \\
    Encoder entropy coefficient & $\entropycoef$ & 5.0 \\
    Consistency loss coefficient & $\consistencycoef$ & 0.01 \\
    Reward coefficient & $\rewardcoef$ & 10.0 \\
    Discount coefficient & $\discountcoef$ & 50.0 \\
    Actor entropy coefficient & $\actorentropycoef$ & 0.01 \\
    Actor entropy threshold & $\actorentropythreshold$ & 0.1 \\
    \midrule
    Environment steps & --- & $100$K \\
    Frame skip & --- & $4$ \\
    Frame down-sampling & --- & $64 \times 64$ \\
    Frame gray-scaling & --- & Yes \\
    Frame stack & --- & $4$ \\
    Terminate on live loss & --- & Yes \\
    Max frames per episode & --- & $108$K \\
    Max no-ops & --- & $30$ \\
    \midrule
    Observation learning rate & --- & $0.0001$ \\
    Dynamics learning rate & --- & $0.0001$ \\
    Actor learning rate & --- & $0.0001$ \\
    Critic learning rate & --- & $0.00001$ \\
    \midrule
    Transformer embedding size & --- & $256$ \\
    Transformer layers & --- & $10$ \\
    Transformer heads & --- & $4 \times 64$ \\
    Transformer feedforward size & --- & $1024$ \\
    Latent state predictor units & --- & $4 \times 512$ \\
    Reward predictor units & --- & $4 \times 256$ \\
    Discount predictor units & --- & $4 \times 256$ \\
    Actor units & --- & $4 \times 512$ \\
    Critic units & --- & $4 \times 512$ \\
    Activation function & --- & SiLU \\
    \bottomrule
  \end{tabular}
\end{table}

\begin{table}
  \centering
  \caption{Number of parameters of our models.}
  \label{tab:model-size}
  \footnotesize
  \begin{tabular}{lcr}
    \toprule
    Model & Symbol & \# Parameters \\
    \midrule
    Observation model & $\obsparam$ & $8.2$M \\
    Dynamics model & $\dynparam$ & $10.8$M \\
    Actor & $\actorparam$ & $1.3$M \\
    Critic & $\criticparam$ & $1.3$M \\
    \midrule
    World model & --- & $19$M \\
    Actor-critic & --- & $2.6$M \\
    \midrule
    Total & --- & $21.6$M \\
    \midrule
    \begin{tabular}{@{}l@{}}Encoder + actor \\ (at inference time)\end{tabular}
    & --- & $4.4$M \\
    \bottomrule
  \end{tabular}
\end{table}

\end{document}